\documentclass[10pt,journal,compsoc]{IEEEtran}
\IEEEoverridecommandlockouts

\usepackage{amsmath,amsthm,amsfonts,amssymb}
\usepackage{cite}
\usepackage{bm}
\usepackage{bbm}
\usepackage{url}
\usepackage{array}
\usepackage{color}
\usepackage{multirow}
\usepackage{booktabs}
\usepackage[table,xcdraw]{xcolor}
\usepackage{enumerate}
\usepackage{enumitem}
\usepackage{graphicx,subcaption}

\usepackage[english]{babel}
\usepackage{algorithm}
\usepackage[noend]{algpseudocode}
\usepackage{setspace}
\usepackage{makecell}

\theoremstyle{plain}

\newtheorem{rem}{Remark}
\newtheorem{sty1}{Theorem}
\newtheorem{defi}[sty1]{Definition}

\usepackage[english]{babel}
\usepackage{algorithm}
\usepackage[noend]{algpseudocode}

\allowdisplaybreaks[4]

\newcommand{\RR}{I\!\!R}

\makeatletter
\newcommand{\ssymbol}[1]{^{\@fnsymbol{#1}}}
\newcommand{\algmargin}{\the\ALG@thistlm}
\makeatother

\newlength{\whilewidth}
\algdef{SE}[parWHILE]{parWhile}{EndparWhile}[1]
  {\parbox[t]{\dimexpr\linewidth-\algmargin}{%
     \hangindent\whilewidth\strut\algorithmicwhile\ #1\ \algorithmicdo\strut}}{\algorithmicend\ \algorithmicwhile}%
\algnewcommand{\parState}[1]{\State%
  \parbox[t]{\dimexpr\linewidth-\algmargin}{\strut #1\strut}}

\begin{document}
\title{Learning-based Autonomous Channel Access in the Presence of Hidden Terminals}

\author{
Yulin~Shao,~\IEEEmembership{Member,~IEEE},
Yucheng~Cai,
Taotao~Wang,~\IEEEmembership{Member,~IEEE},
Ziyang~Guo,~\IEEEmembership{Member,~IEEE},
Peng~Liu,~\IEEEmembership{Member,~IEEE},
Jiajun~Luo,
Deniz~G{\"u}nd{\"u}z,~\IEEEmembership{Fellow,~IEEE}

\thanks{Y. Shao and D. G{\"u}nd{\"u}z are with the Department of Electrical and Electronic Engineering, Imperial College London, London SW7 2AZ, U.K. (e-mail: \{y.shao,~d.gunduz\}@imperial.ac.uk).

\noindent Y. Cai and T. Wang are with the College of Electronic and Information Engineering, Shenzhen University, Shenzhen, China  (e-mail: \{caiyucheng,~ttwang\}@szu.edu.cn).

\noindent Z. Guo, P. Liu, and J. Luo are with the Wireless Technology Lab, 2012 Labs, Huawei, China (e-mail:\{guoziyang,~jeremy.liupeng,~luojiajun4\}@huawei.com).
}
}

\IEEEtitleabstractindextext{%
\begin{abstract}
We consider the problem of autonomous channel access (AutoCA), where a group of terminals try to discover a communication strategy with an access point (AP) via a common wireless channel in a distributed fashion. Due to the irregular topology  and the limited communication range of terminals, a practical challenge for AutoCA is the hidden terminal problem, which is notorious in wireless networks for deteriorating the throughput and delay performances. To meet the challenge, this paper presents a new multi-agent deep reinforcement learning paradigm, dubbed MADRL-HT, tailored for AutoCA in the presence of hidden terminals. MADRL-HT exploits topological insights and transforms the observation space of each terminal into a scalable form independent of the number of terminals. To compensate for the partial observability, we put forth a look-back mechanism such that the terminals can infer behaviors of their hidden terminals from the carrier sensed channel states as well as feedback from the AP. A window-based global reward function is proposed, whereby the terminals are instructed to maximize the system throughput while balancing the terminals' transmission opportunities over the course of learning. Extensive numerical experiments verified the superior performance of our solution benchmarked against the legacy carrier-sense multiple access with collision avoidance (CSMA/CA) protocol.
\end{abstract}

\begin{IEEEkeywords}
Multiple channel access, hidden terminal, multi-agent deep reinforcement learning, Wi-Fi, proximal policy optimization.
\end{IEEEkeywords}}

\maketitle
\IEEEdisplaynontitleabstractindextext
\IEEEpeerreviewmaketitle

\section{Introduction}\label{sec:I}
Autonomous channel access (AutoCA) is a class of multiple-access control (MAC) problems, where a group of terminals communicate with an access point (AP) on a shared channel in a decentralized fashion \cite{CSMA,yu1,guo,LTE,CR1,opt1,POMDP}.
Due to the multiple-access interference, at any time only one terminal can communicate with the AP;
when two or more terminals transmit simultaneously,
a collision happens and the transmissions fail.
In AutoCA, the goal of the terminals is to discover a set of transmission policies autonomously such that the system throughput is maximized and the fairness among terminals is guaranteed. AutoCA is essential to communication networks that lack well-governed infrastructures, such as communications systems operated on the unlicensed band \cite{CSMA2,Zigbee,V2X}, to provide high bandwidth efficiency and low-latency services.

The classical solution to the AutoCA problem is human-crafted random access policies. A typical example is the contention-based IEEE 802.11 carrier-sense multiple access with collision avoidance (CSMA/CA) \cite{CSMA,CSMA2}, which is now widely used in Wi-Fi networks. When operated with CSMA/CA, distributed terminals compete with each other for channel access opportunities: each terminal carrier senses the channel before transmission (i.e., the listen-before-talk (LBT) protocol); and retransmits after a binary exponential back-off (BEB) in the case of transmission failures, i.e., the average back-off time is doubled after a transmission failure to reduce the collision probability.

Recent advances in deep reinforcement learning multiple access (DRLMA) open up a new perspective for solving the AutoCA problem \cite{opt1,opt2,yu1,yu2,yu3,LTE,guo,CR1,CR3,CR4,CR5,CR6}. With DRLMA, distributed terminals endowed with intelligence learn to cooperate, as opposed to competing, to share the spectrum harmonically. In particular, the group of terminals are expected to adapt to each other's transmission policies to avoid collisions and balance channel-access opportunities. To this end, the AutoCA problem is formulated as either a Markov decision process (MDP) \cite{opt1,opt2,yu1,LTE,Significant2020,CR6} or a partially observable MDP (POMDP) \cite{yu2,yu3,guo,CR1,CR3,CR5}, depending on whether the environment is fully observable to the terminals or not. The MDP or POMDP is then solved by DRL or multi-agent DRL (MADRL), wherein terminals progressively learn a good set of policies from their past transmission history, carrier sensed behaviors of other terminals, and additional information fed back from the AP/channel. Compared with legacy random access policies, DRLMA often exhibits much better performance in terms of throughput, delay, and jitter, demonstrating the great potential to solve the AutoCA problem in a variety of network scenarios.

A notorious problem in wireless networks is the hidden terminal problem \cite{Hidden,Hidden2}. Hidden terminals are terminals that are out of each other's communication range.
Hidden terminals cannot detect each other's transmission behaviors via carrier sensing, and hence, their transmissions to the common receiver can collide, resulting in a waste of spectrum.
In Wi-Fi networks, for example, the LBT protocol breaks down when there are hidden terminals, due to the inaccurate carrier-sensed channel states; and the throughput and delay performances deteriorate significantly \cite{Hidden2}.
To alleviate this problem, the IEEE 802.11 standard introduced the request to send/clear to send (RTS/CTS) mechanism. However, RTS/CTS falls short of an ideal protocol as it incurs additional handshake overhead and new problems such as hidden receiver and exposed receiver \cite{RTSCTS1,RTSCTS2} -- as a consequence, it can even deteriorate the performance of CSMA/CA \cite{RTSCTS2}, especially in machine-type communications with short packet lengths. Nowadays, the RTS/CTS mechanism is deactivated in most Wi-Fi devices.\footnote{In commercial products, the RTS/CTS mechanism is activated only when the packet size is larger than an RTS threshold. Often, the default and recommended settings of the packet size and RTS threshold disable the RTS/CTS function. Examples include TP-Link TL-WR841N Series, Huawei AR100 Series, Cisco ISA500 Series, to name a few.} On the other hand, prior works on DRLMA focused exclusively on the ideal case, where all the terminals are audible to each other and the hidden terminal problem is completely omitted. In the practical setup with hidden terminals, a terminal can observe inaccurate channel states; its transmissions can collide with the transmissions of a hidden terminal that it has never seen. Worse still, the transmission policy of the hidden terminal varies across time, yielding non-stationary hidden dynamics to the terminal. In this context, designing a DRLMA algorithm to tackle the impact of hidden terminals is an important and challenging task.

To fill this gap, this paper presents MADRL-HT, a partially observable MADRL paradigm tailored for AutoCA in the presence of hidden terminals.
Three innovations that underpin the design of MADRL-HT are terminal clustering, the look-back mechanism, and a window-based reward function.
\begin{enumerate}
\item \textbf{Terminal clustering}. For individual terminal, we put forth a clustering method to partition all other terminals according to their number of communication hops to the considered terminal. In so doing, the environment faced by each terminal can be summarized as the dynamics of two consolidated groups of terminals, thereby simplifying the observation space by a large margin.
\item \textbf{Look-back mechanism}. To compensate for the partial observability and alleviate the hidden non-stationarity, we proposed a look-back mechanism to revise the carrier-sensed channel states and infer the hidden terminals' behaviors via the acknowledgment (ACK) of the AP.
\item \textbf{Window-based reward}. Unlike prior works where the reward is often designed to be event-triggered, we put forth a new reward function based on the relative number of successful transmissions in a look-back window of time. Our reward motivates the agents to avoid collision and maximize the network throughput, while balancing the successful transmissions among terminals to ensure fairness.
\end{enumerate}

Extensive numerical experiments verify the superior performance of our MADRL-HT solution in various topologies.
Compared with the legacy CSMA/CA protocol, the proposed MADRL-HT solution exhibits remarkable performance gains: the $\alpha$-fairness is improved by up to $46\%$; the  average packet collision rate (PCR) is reduced by up to $37\%$; the average packet delay is reduced by $99.2\%$, and the delay jitter is reduced by $99.9\%$.

The remainder of this paper is organized as follows.
Section \ref{sec:review} reviews the prior arts on AutoCA and DRLMA.
Section \ref{sec:II} formulates the AutoCA problem with hidden terminals;
Section \ref{sec:III} analyzes the AutoCA problem and demonstrates how to partition the terminals;
Section \ref{sec:IV} presents our MADRL-HT solution, including the state and reward designs tailored for AutoCA with hidden terminals;
Section \ref{sec:V} presents the experimental results to verify the superiority of our MADRL-HT solution;
Section \ref{sec:Conclusion} concludes this paper.

\section{Related Work}\label{sec:review}
The standard solution in IEEE 802.11 to the AutoCA problem is CSMA/CA with binary exponential back-off (BEB) \cite{CSMA,CSMA2}. To be more specific, the back-off time of CSMA/CA is randomly sampled from $[0,\text{CW}]$, where $\text{CW}$ is the length of a contention window. In the case of a collision, $\text{CW}$ will be doubled to reduce the probability of contention. The setting of $\text{CW}$ has a significant impact on the efficiency of CSMA/CA. As BEB is a heuristic scheme, several recent works \cite{opt1,opt2} employed DRL to learn better policies for setting $\text{CW}$ under different network conditions, whereby the system throughput is optimized. 

As opposed to optimizing the parameters on top of CSMA/CA, a more promising and interesting method that exploits the full potential of DRL is learning AutoCA protocols from scratch \cite{yu1,yu2,yu3,LTE,guo}. In this line of work, early studies \cite{yu1,yu2,LTE} focused on the heterogeneous setup where an intelligent terminal (or a cluster of terminals with a central controller) learns to coexist with other terminals operated with legacy MAC protocols, such as Aloha, TDMA, CSMA/CA, and LTE. These works formulated single-agent RL problems, in which an intelligent agent learns to adapt to other non-intelligent agents' unknown policies from its observations. Compared with MARL, single-agent RL is less challenging as the environment faced by the agent is stationary.

Motivated by these research efforts, two recent works \cite{yu3,guo} made further advances on the AutoML problem. In \cite{yu3}, the authors generalized \cite{yu2} to a multi-agent setup, where multiple intelligent terminals make independent transmission decisions in the heterogeneous network. The unreliable physical channel is also taken into account in \cite{yu3}.
Ref. \cite{guo}, on the other hand, considered a homogeneous setup where all terminals are intelligent and adapted the well-known MARL algorithm, Qmix \cite{qmix}, to solve the AutoCA problem.
In \cite{yu2,yu3,guo}, the intelligent terminals are endowed with carrier sensing capabilities, and they can utilize the feedback from the AP to make transmission decisions. 
Another line of work \cite{Hoydis1,Hoydis2} studied the joint learning of MAC signaling and access protocols with MADRL.

A shortcoming of the above-mentioned works is that the hidden terminal problem is completely omitted. By carrier sensing, the channel occupation is fully observable to the agents. In a more practical setup with hidden terminals, a terminal can observe inaccurate channel states; its transmissions can collide with the transmissions of a hidden terminal that it has never seen. Worse still, the transmission policy of the hidden terminal varies across time, yielding non-stationary hidden dynamics to each terminal. In this context, designing a DRLMA algorithm to tackle the impact of hidden terminals is a challenging task, which we aim to address in this paper.

It is worth noting that another line of work that addresses the multiple-access problems with DRL is dynamic spectrum access (DSA) in cognitive radios \cite{CR1,CR3,CR4,CR5,CR6}. Under their framework, there are two classes of users in the network: primary users (PUs) and secondary users (SUs). The spectrum is licensed to the PUs, and hence, they have the highest channel access priorities. In an effort to maximize the spectrum efficiency, the SUs are allowed to access the unused spectrum provided that they do not harm the channel usages of PUs. To this end, the SUs learn to identify the unused spectrum, predict the channel-access patterns of the PUs, and cooperate harmonically with each other. As can be seen, DSA with cognitive radios can be viewed as an AutoCA problem under the heterogeneous setup with a class of prioritized PUs.

\section{Problem Statement}\label{sec:II}
\begin{figure}[t]
\centering
\includegraphics[width=1\linewidth]{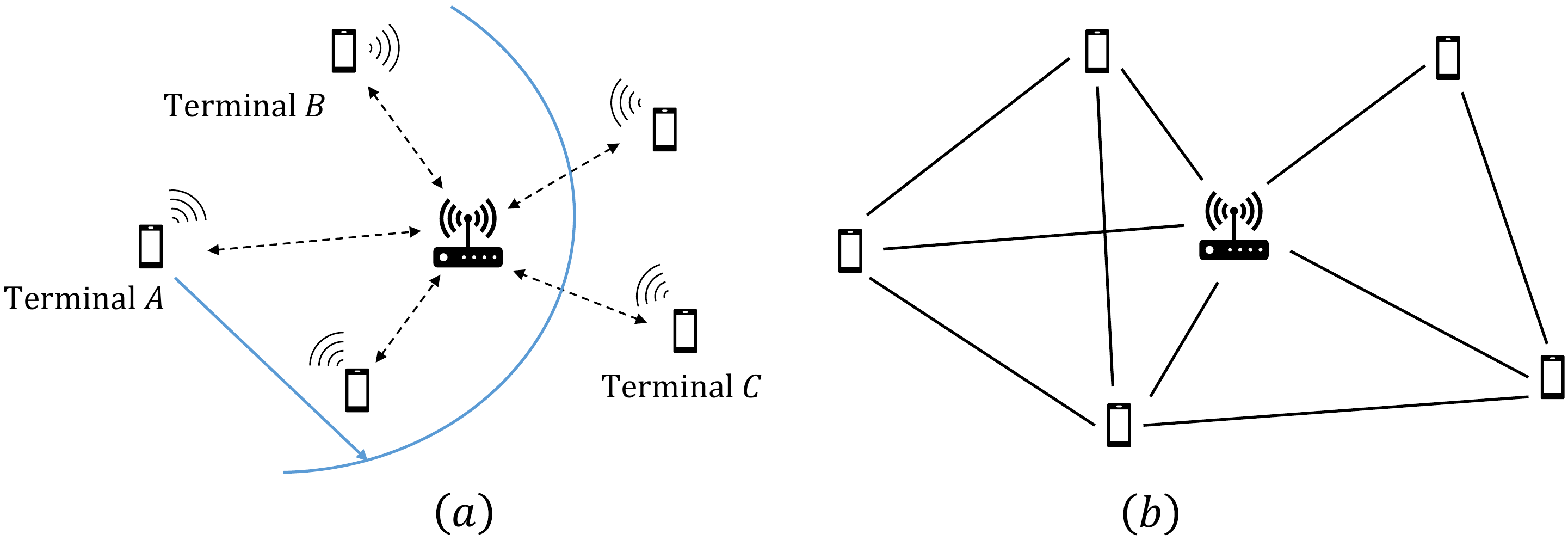}
\caption{A Wi-Fi BSS with one AP and $N$ terminals sharing a common uplink channel: (a) the physical network; (b) the graph whose vertices represent the terminals and the edges connect terminals that are within each other's communication range.}
\label{fig:model1}
\end{figure}

We consider a conventional Wi-Fi basic service set (BSS) with one AP and $N$ terminals formed in a star topology, as shown in Fig. \ref{fig:model1}(a). 
The terminals are indexed by $n=1,2,...,N$.
We focus on the uplink transmissions from the terminals to the AP. All terminals share the same uplink frequency spectrum, which are referred to as the common channel. Time is divided into slots. We consider the saturated traffic setup, i.e., the transmission buffers of the terminals are nonempty in any time slot, and a terminal has a packet to transmit whenever there is an idle slot.
Throughout the paper, we assume that a transmission failure can only be caused by collisions, but not the quality of the underlying physical channel.

Not all terminals are within each other's communication range. In Fig. \ref{fig:model1}(a), for example, terminals A and B are within each other's communication range. Therefore, terminal A can sense terminal B's behaviors (i.e., transmit or not) when A is idle, and vice versa. On the other hand, terminals A and C are out of each other's communication range, and hence, they cannot infer each other's transmissions from carrier sense -- they are ``hidden terminals'' to each other.

In the presence of hidden terminals, this paper aims to address the AutoCA problem, i.e., how a group of agents can reach a multiple-access protocol autonomously to  maximize the network throughput while guaranteeing fairness among each other. More rigorously,  we formulate the AutoCA problem using the $\alpha$-fairness measure in the following.

Let us denote the set of channel-access policies of the $N$ terminals by $\bm{\pi}=\{\pi_n:n=1,2,\ldots,N\}$, where $\pi_n$ is the policy of the $n$-th terminal.
When operated with $\bm{\pi}$, we denote by  $\mathcal{T}_n(\bm{\pi})$ the throughput achieved by the $n$-th terminal. In particular, $\mathcal{T}_n(\bm{\pi})$ can be written as 
\begin{equation}\label{eq:throughput}
    \mathcal{T}_n(\bm{\pi})\triangleq \frac{M_n(\bm{\pi},T)D}{T},
\end{equation}
where $D$ is the packet length measured in terms of number of slots and we assume all terminals have the same packet length;  and $M_n(\bm{\pi},T)$ is the number of packets successfully transmitted by the $n$-th terminal in a period of $T$ slots under the set of policies $\bm{\pi}$. The $\alpha$-fairness value of the $n$-th terminal is then given by \cite{alpha}
\begin{equation}\label{afairness}
    f\big(\mathcal{T}_n(\bm{\pi})\big)=\begin{cases}
        \log\big(\mathcal{T}_n(\bm{\pi})\big), & \alpha=1,\\
        \frac{\big(\mathcal{T}_n(\bm{\pi})\big)^{1-\alpha}}{(1-\alpha)}, & \alpha\neq 1.
    \end{cases}
\end{equation}

In AutoCA, the goal of the terminals is to find the optimal set of channel-access policies $\bm{\pi}^*$ in a decentralized manner to maximize the $\alpha$-fairness function $F(\bm{\pi})$, which is the sum of all $\alpha$-fairness values. Specifically, we have
\begin{equation}\label{goalpi}
    \bm{\pi}^*=\arg\max_{\bm{\pi}} \sum_{n=1}^{N}f\big(\mathcal{T}_n(\bm{\pi})\big)\triangleq\arg\max_{\bm{\pi}} F(\bm{\pi}).
\end{equation}
With the $\alpha$-fairness measure, the optimal set of chan\-nel-access policies $\bm{\pi}^*$ is the policy set that maximizes the total throughput of the Wi-Fi BSS while balancing the throughput of individual terminals equally.

This paper explores an MADRL solution to the AutoCA problem, where individual terminals in the BSS learn to optimize their transmission policies based on their sensing and perception of the environment. 
In particular, under the IEEE 802.11 standards, two pieces of information that a terminal can exploit to learn the optimal policy are:
\begin{enumerate}
    \item The carrier-sensed channel states. When it is idle, a terminal can actively sense the occupancy of the common channel within its communication range, and hence, infer and adapt to other terminals' transmission behaviors. There are two possible carrier-sensed channel states: ``idle'' and ``busy''.
    \item Feedback from the AP. The AP broadcasts an AC\-K/NACK signal (NACK means that the a collision happens and no ACK is received at the end of transmission) following every transmission, indicating whether a packet has been received successfully or not. We assume that this $1$-bit feedback from the AP is received by all the terminals reliably. 
\end{enumerate}



\begin{rem}
It is worth noting that packet length is conventionally defined in terms of the number of bytes, such as the maximum transmission unit (MTU). Here, we define the packet length in the manner of transmit opportunity (TXOP) \cite{TXOP,FLOAC}: the time duration for which a terminal can transmit after it has gained the channel.
That is, when a terminal decides to transmit, it occupies the channel for $D$ slots until the whole packet is transmitted.
\end{rem}

\section{Terminal Clustering}\label{sec:III}
For any terminal in the Wi-Fi BSS, the environment it faces consists of the other $N-1$ terminals. 
To develop a sound transmission policy that coexists well with others, the first step is to infer each of the other terminals' transmission policies.
The partial observability of our problem, however, makes the inference an elusive task since it is impossible to extrapolate each of the other $N-1$ terminals' transmission histories from the limited carrier-sensed channel states and AP's feedback.

To address the above challenge, this section puts forth a terminal clustering method: for each terminal in the BSS, we will show how to partition the other $N-1$ terminals such that their transmission histories can be summarized as the histories of two consolidated terminals, thereby yielding a compact observation space for each terminal in the BSS.
Along with the discussion, we will also define some basic terminologies to be used in the rest of this paper.

To start with, we study the topology of a Wi-Fi BSS by treating it as an undirected graph -- its vertices correspond to the terminals and the AP; and there is an edge connecting two vertices if and only if the corresponding terminals are within each other's communication range. As an example, the Wi-Fi BSS in Fig. \ref{fig:model1}(a) can be transformed to the graph shown in Fig. \ref{fig:model1}(b).

\begin{defi}[$n$-hop neighbor]\label{defi:1}
Consider two vertices A and B on a graph. Vertex B is an $\ell$-hop neighbor of vertex A if and only if vertex A can reach vertex B in at least $\ell$ hops.
\end{defi}

Given Definition \ref{defi:1}, we have two immediate results from Fig.~\ref{fig:model1}(b), thanks to the star topology of the BSS.
\begin{itemize}
    \item In the Wi-Fi BSS, all terminals are one-hop (OH) neighbors to the AP.
    \item For a terminal A in the BSS, any other terminal is either a OH or a two-hop (TH) neighbor of terminal A.
\end{itemize}

\begin{figure}[t]
\centering
\includegraphics[width=1\linewidth]{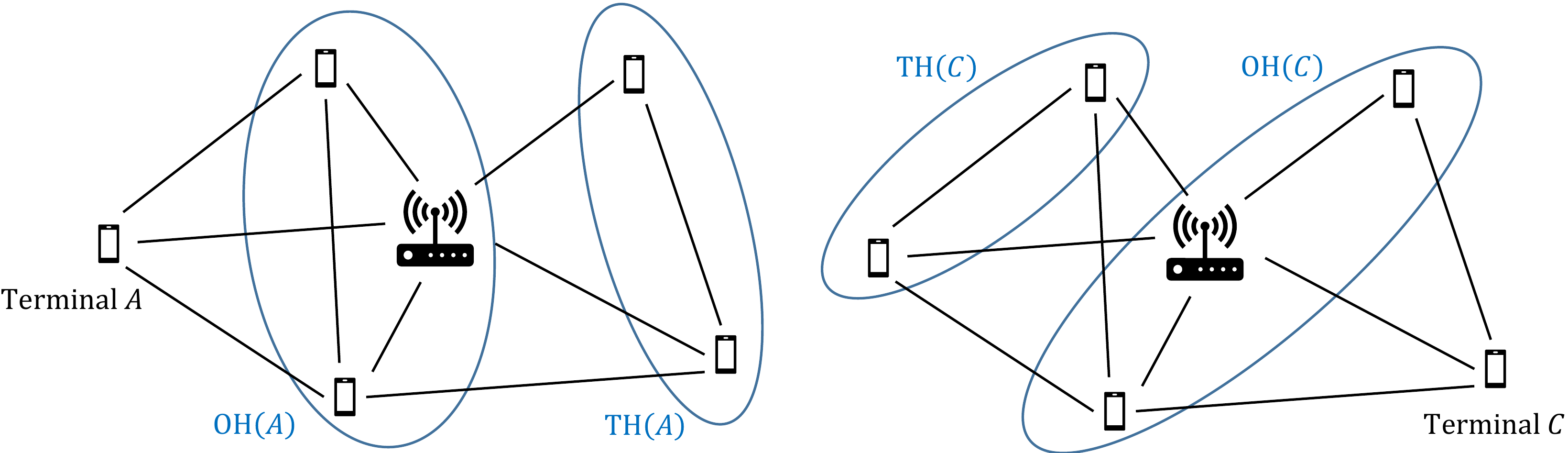}
\caption{Any terminal in the Wi-Fi BSS has only OH and TH neighbors. Different terminals observe different sets of OH and TH neighbors.}
\label{fig:3.1}
\end{figure}

In light of the above, for the $n$-th terminal in the Wi-Fi BSS, we can partition the other $N-1$ terminals into two sets: one consists of all OH neighbors of the $n$-th terminal, denoted by OH($n$), and the other consists of all TH neighbors of the $n$-th terminal, denoted by TH($n$). An example is given in Fig.~\ref{fig:3.1}, in which any other terminal in the BSS is either an OH or a TH neighbor to terminal A or C. We emphasize that different terminals have different sets of OH and TH neighbors.

Another key observation is that, for the $n$-th terminal, all the terminals in OH($n$) can be viewed as a single consolidated terminal because 1) the $n$-th terminal senses them in the same way; and 2) their actions (transmit or not) have the same effects on the $n$-th terminal. Likewise, all the terminals in TH($n$) can also be viewed as a single consolidated terminal. Therefore, from the perspective of the $n$-th terminal, the transmission actions of all other terminals can be summarized as the transmission actions of  two consolidated terminals: OH($n$) and TH($n$).

\begin{defi}[Actions of OH and TH neighbors]
Suppose there are $N$ terminals in a Wi-Fi BSS. The action of the $n$-th terminal in slot $t$, denoted by $a_n^t\in\{0,1\}$, indicates whether the $n$-terminal is transmitting $(a_n^t=1)$ or idle $(a_n^t=0)$ in slot $t$.

Suppose the $n$-th terminal has $M-1$ OH neighbors and $N-M$ TH neighbors. Let $\text{OH}(n)\triangleq\{i_1,i_2,\ldots,i_{M-1}\}$ and $\text{TH}(n)\triangleq\{j_1,j_2,\ldots,j_{N-M}\}$. We define
\begin{eqnarray*}
a_{\text{OH}(n)}^t
\hspace{-0.2cm}&\triangleq&\hspace{-0.2cm}
a_{i_1}^t\vee a_{i_2}^t\vee\cdots\vee a_{i_{M-1}}^t, \\
a_{\text{TH}(n)}^t
\hspace{-0.2cm}&\triangleq&\hspace{-0.2cm}
a_{j_1}^t\vee a_{j_2}^t\vee\cdots\vee a_{j_{N-M}}^t,
\end{eqnarray*}
to indicate, respectively, whether an OH and a TH neighbor of the $n$-th terminal transmits in slot $t$ or not,  where $\vee$ stands for the OR operation.
\end{defi}

\begin{rem}[Carrier sensing]
The $n$-th terminal carrier senses the channel as long as $a_n^t=0$.
Two outputs of carrier sensing, ``idle'' and ``busy'', correspond to $a_{\text{OH}(n)}^t=0$ and $a_{\text{OH}(n)}^t=1$, respectively.

\end{rem}

\begin{rem}[Listen-before-talk]
To be compatible with legacy IEEE 802.11 MAC standards, we impose the listen-before-talk (LBT) constraint on terminals' actions. Specifically, a terminal tries to access the channel only after it sensed that the channel has been idle for at least a distributed coordination function interframe spacing (DIFS). As a result, there will be at least DIFS idle slots between consecutive transmissions of a single terminal or two terminals that are within OH.
\end{rem}

\section{An MADRL Solution}\label{sec:IV}
Based on our discussions in Sections \ref{sec:II} and \ref{sec:III}, this section formulates AutoCA as a multi-agent reinforcement learning (MARL) problem and presents MADRL-HT, an  MADRL solution tailored for AutoCA in the presence of hidden terminals.
In MADRL-HT, terminals interact and negotiate autonomously with each other to reach a multiple-access protocol.
During the interactions, each terminal infers other terminals' behaviors (summarized as two consolidated terminals) from the carrier-sensed channel states and AP's feedback. Transmission decisions are then made to adapt to other terminals' policies.

\begin{figure}[t]
\centering
\includegraphics[width=0.7\linewidth]{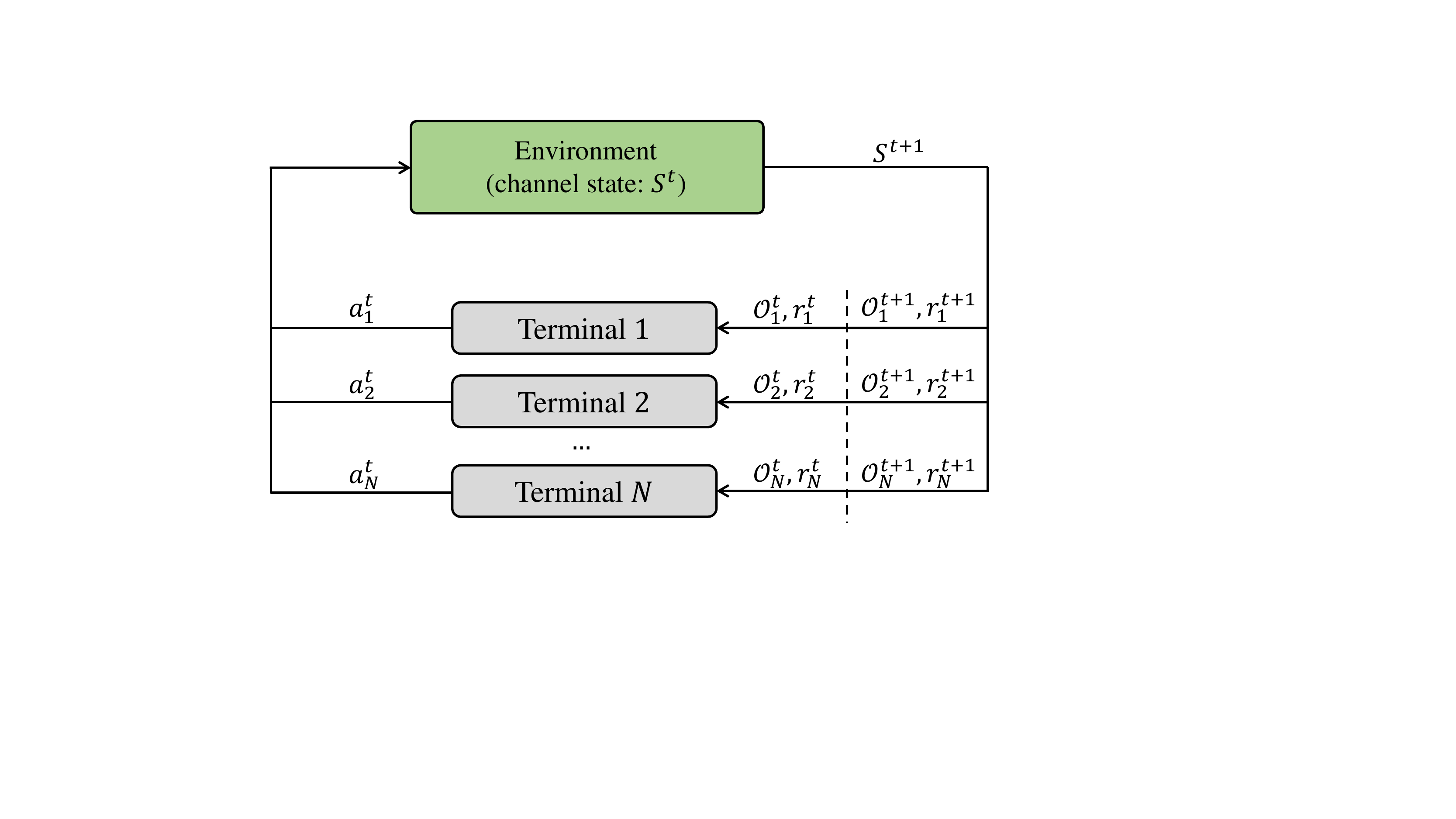}
\caption{Modelling the AutoCA problem with hidden terminals under the framework of MARL.}
\label{fig:1}
\end{figure}
A general MARL problem \cite{qmix,MARL} can be defined as a tuple $(\mathcal{N},\allowbreak\mathcal{S},\allowbreak\{\mathcal{O}_n\}_{n\in\mathcal{N}},\allowbreak\mathcal{A},\allowbreak\{r_n\}_{n\in\mathcal{N}})$, where $\mathcal{N}=\{1,2,...,N\}$ denotes the set of $N$ agents; $\mathcal{S}$ denotes the state space of the environment that the agents reside in; $\mathcal{O}_n$ denotes the observation space of the $n$-th agent; $\mathcal{A}$ denotes the action space of each agent; $r_n:\mathcal{S}\times\mathcal{A}\times\mathcal{S}\to \RR$ denotes the immediate reward function of the $n$-th agent, which defines the reward that the environment assigns to an agent.
In the above context, the MARL problem associated with AutoCA is illustrated in Fig. \ref{fig:1}. At any time step $t$, the action of the $n$-th terminal is defined as $a_n^t\in \mathcal{A}=\{'\text{transmit}','\text{idle}'\}$; the state of the environment $S^t\in\mathcal{S}$ represents the channel usages and is the collection of actions of all the terminals. Each terminal has a local observation of $S^t$ denoted by $\mathcal{O}^t_n$. In particular, $\mathcal{O}^t_n$ is determined by the carrier sensed channel state of the $n$-th terminal and the ACK feedback from the AP. Based on the observation $\mathcal{O}^t_n$ and the rewards received, each agent takes an action $a^t_n$. The set of actions $\{a_n^t \}_{n\in\mathcal{N}}$ steers the environment moving to a new state $S^{t+1}$ and a reward $r^t_n$ is assigned to each terminal to evaluate its last action. Over the course of MARL, the terminals aim to discover a set of transmission policies in a distributed manner -- each terminal learns from its interactions with other terminals, its partially observed channel states, and the rewards it receives -- to avoid collisions and maximize the BSS throughput, while guaranteeing fairness among each other.

\subsection{Partial observability}
As stated above, the environment faced by a terminal in the Wi-Fi BSS is the other $N-1$ terminals. At any slot $t$, a decision (transmit or not) has to be made by each terminal based on the past behaviors of all terminals. In this context, we define the global state of the BSS at time slot $t$ as the past actions of the terminals.

\begin{defi}[Global state]
The global state of a Wi-Fi BSS at slot $t$, denoted by $S^t$, is the transmission history of all the terminals in the last $W$ slots:
\begin{equation}\label{eq:state}
    S^t\triangleq\big\{a_n^\tau:n=1,2,...,N,\tau=t-1,t-2,\ldots,t-W\big\},
\end{equation}
where $W$ is the length of a look-back window, reflecting how many slots we look back into the past to make the transmission decision in slot $t$. The look-back window is necessary in that the terminals' behaviors can be  non-Markovian in general, and hence, a decision has to be made based on a period of past behaviors such that the observed environment is approximately Markovian. Typically, $W$ should be much larger than the packet length $D$.
\end{defi}

The global state, however, is only partially observable to each terminal, as a terminal has only a local view/observation of other terminals' behaviors. Specifically, the observation of a terminal is subject to the following two constraints:\footnote{In practice, there can be more constraints, such that the power saving demands of terminals.}
\begin{itemize}
    \item Constraint 1: A terminal cannot sense the channel while transmitting.
    \item Constraint 2: Hidden terminals are inaudible to each other.
\end{itemize}

Therefore, it is important to characterize the discrepancies between a terminal's local observations and the global state $S^t$.

Consider the $n$-th terminal. $a_{\text{OH}(n)}^t$ is observable if and only if $a_n^t=0$, due to Constraint 1. On the other hand, $a_{\text{TH}(n)}^t$ is unobservable with carrier sensing. Nevertheless, notice that a successful transmission is always followed by an ACK, which is broadcasted by the AP to all the terminals in the BSS. This suggests that a terminal can further infer the behaviors of other terminals, including both its OH and TH neighbors, from the feedback it receives from the AP.

\begin{defi}[Partial observations of a terminal]
In a Wi-Fi BSS, the observation of the $n$-th terminal at the beginning of a slot $t$ is defined as
\begin{equation}\label{eq:obs}
    \mathcal{O}_n^t\triangleq\big\{a_n^\tau,o_{\text{OH}(n)}^\tau,o_{\text{TH}(n)}^\tau:\tau=t-1,t-2,\ldots,t-W\big\}
\end{equation}
where $o_{\text{OH}(n)}^t$ and $o_{\text{TH}(n)}^t$ are estimates of $a_{\text{OH}(n)}^t$ and $a_{\text{TH}(n)}^t$, respectively, and are inferred by the $n$-th terminal from its carrier sensed channel states and the feedback from the AP. 
\end{defi}

\begin{algorithm}[t]
\caption{The look-back mechanism to infer the transmission histories of $\text{OH}(n)$ and $\text{TH}(n)$.}
\label{algo:observations}
\begin{algorithmic}[1]
\State{\textbf{Input:} AP's feedback at slot $t$, i.e., ACK or NACK.}
\State{\textbf{Output:} Estimated transmission history $o_{\text{OH}(n)}^{\tau}$ and $o_{\text{TH}(n)}^{\tau}$, where $\tau\in \Omega= \{t-D+1,...,t-1,t\}$.}
\State{\underline{\textbf{Carrier sensing}}}
\For{$\tau\in\Omega$}
    \State $o_{\text{TH}(n)}^{\tau}=\text{unk}$.
    \If {$a_n^{\tau}=0$}
        $o_{\text{OH}(n)}^{\tau}=a_{\text{OH}(n)}^{\tau}$;
    \Else~
        $o_{\text{OH}(n)}^{\tau}=\text{unk}$.
    \EndIf
\EndFor
\State{\underline{\textbf{Look-back revision}}}
\If {ACK is received at slot $t$}
    \If {$a_n^{\tau}=1$, $\forall \tau\in\Omega$}
        \State $o_{\text{OH}(n)}^{\tau\in\Omega}=0,o_{\text{TH}(n)}^{\tau\in\Omega}=0$;
    \ElsIf {$a_n^{\tau}=0$ and $o_{\text{OH}(n)}^\tau=0$, $\forall \tau\in\Omega$}
        \State $o_{\text{TH}(n)}^{\tau\in\Omega}=1$;
    \ElsIf {$a_n^{\tau}=0$ and $o_{\text{OH}(n)}^\tau=1$, $\forall \tau\in\Omega$}
        \State $o_{\text{TH}(n)}^{\tau\in\Omega}=0$.
    \EndIf
\EndIf
\end{algorithmic}
\end{algorithm}

In MADRL-HT, we put forth a look-back mechanism to determine $o_{\text{OH}(n)}^{\tau}$ and $o_{\text{TH}(n)}^{\tau}$ for the $n$-th terminal, $n=1,2,...,N$.
The look-back mechanism is summarized in Algorithm \ref{algo:observations} and consists of two steps.
At any time slot $t$, we feed the AP's feedback, i.e., ACK or NACK, into Algorithm \ref{algo:observations} and estimate the transmission histories of the $n$-th terminal's OH and TH neighbors during the past $D$ slots, i.e., $o_{\text{OH}(n)}^{\tau}$ and $o_{\text{TH}(n)}^{\tau}$, where $\tau\in\Omega\triangleq \{t-D+1,...,t-1,t\}$. Note that $D$ is the packet length measured in terms of the number of slots.

In the first step, the $n$-th terminal estimates the actions of its OH neighbors by carrier sensing:
when $a_{n}^{\tau}=0$, $\tau\in\Omega$, the carrier-sensed output is exactly $a_{\text{OH}(n)}^{\tau}$, hence we set $o_{\text{OH}(n)}^{\tau}=a_{\text{OH}(n)}^{\tau}$; when $a_{n}^{\tau}=1$, the terminal cannot carrier sense and we set $o_{\text{OH}(n)}^{\tau}=\text{unk}$, where $\text{unk}$ is a unique token standing for ``unknown''.
On the other hand, the actions of the TH neighbors are unobservable from carrier sense, hence we simply set  $o_{\text{TH}(n)}^{\tau}=\text{unk}$ for the $n$-th terminal.

The second step is a look-back revision process. At the end of slot $t$, the terminal looks back and revises the carrier-sensed $o_{\text{OH}(n)}^{\tau}$ and $o_{\text{TH}(n)}^{\tau}$ if an ACK is received. Specifically,
\begin{enumerate}
\item if the $n$-th terminal was transmitting in the duration $\Omega$, an ACK at slot $t$ means the transmission is successful and no other terminals were transmitting. Thus, the actions of both OH and TH neighbors are revised to $o_{\text{OH}(n)}^{\tau}=0$, $o_{\text{TH}(n)}^{\tau}=0$, $\forall \tau\in\Omega$.
\item if the $n$-th terminal was idle in the duration $\Omega$, the behaviors of its OH neighbors, $a_{\text{OH}(n)}^\tau$, are fully known via carrier sensing. Therefore, if $o_{\text{OH}(n)}^\tau=a_{\text{OH}(n)}^\tau=0$, i.e., none of its OH neighbors were transmitting during $\Omega$, a received ACK suggests that one of its TH neighbors successfully transmits a packet. We thus revise $o_{\text{TH}(n)}^{\tau}=1$, $\forall \tau\in\Omega$.
\item Finally, if $a_n^{\tau}=0$ and $o_{\text{OH}(n)}^\tau=a_{\text{OH}(n)}^\tau=1$, i.e., at least one of the $n$-th terminal's OH neighbors was transmitting during $\Omega$, the received ACK suggests that there was only one transmitting OH neighbor and all its TH neighbors were idle, because a collision would happen otherwise. Therefore, we revise $o_{\text{TH}(n)}^{\tau}=0$, $\forall \tau\in\Omega$.
\end{enumerate}

\begin{figure}[t]
\centering
\includegraphics[width=0.95\linewidth]{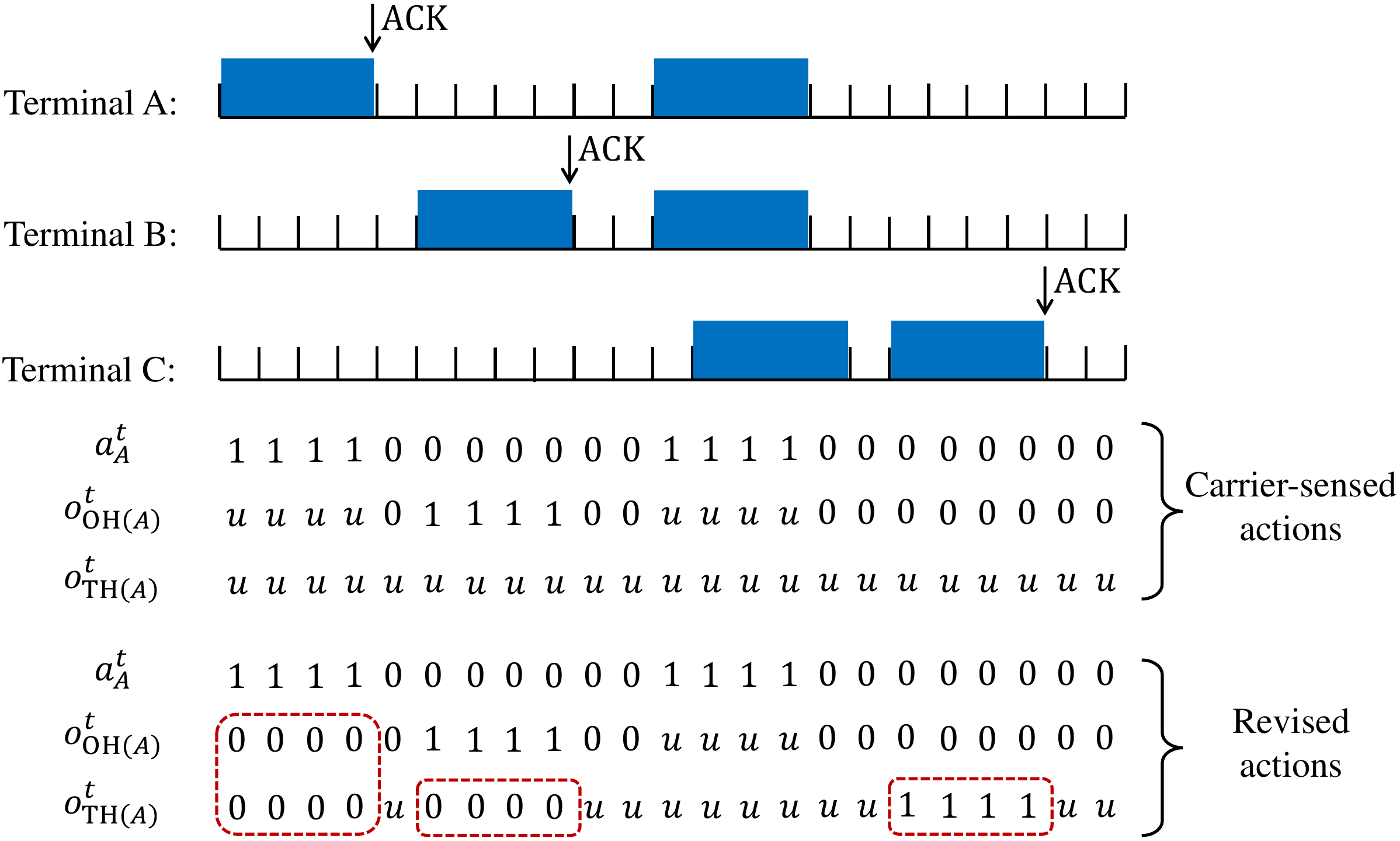}
\caption{An example illustrating how $o_{\text{OH}(A)}^t$ and $o_{\text{TH}(A)}^t$ are determined for a terminal A. We assume there are three terminals A, B, and C in the BSS, where A and B are OH neighbors; A and C are TH neighbors.}
\label{fig:4.1}
\end{figure}

To illustrate the look-back mechanism, an example is given in Fig.~\ref{fig:4.1}, where we consider a group of three terminals A, B, and C in the Wi-Fi BSS: A and B are OH neighbors; A and C are TH neighbors.
Following Algorithm~\ref{algo:observations}, terminal A determines the actions of its OH and TH neighbors in the following way.
First, terminal A estimates $a_{\text{OH}(A)}^t$ via carrier sensing.
As can be seen from the carrier-sensed actions in Fig.~\ref{fig:4.1},
$\{a_A^t\}$ are fully known,
$\{a_{\text{OH}(A)}^t\}$ are known only when $a_A^t=0$ (otherwise, $o_{\text{OH}(A)}^t$ is set to $\text{unk}$);
$\{a_{\text{TH}(A)}^t\}$ are unknown.
Then, terminal A revises its past observations based on the feedback of the AP.
From the first ACK, terminal A knows that none of its OH and TH neighbors were transmitting in the last $D$ slots, $\{o_{\text{OH}(A)}^t\}$ and $\{o_{\text{TH}(A)}^t\}$ are then modified to $0$.
From the second and third ACKs of the AP, terminal A knows that there is a successful transmission from its OH or TH neighbors.
Combined with the carrier sensed channel states, the actions of the TH neighbors can be refined.

\begin{figure}[t]
\centering
\includegraphics[width=0.8\linewidth]{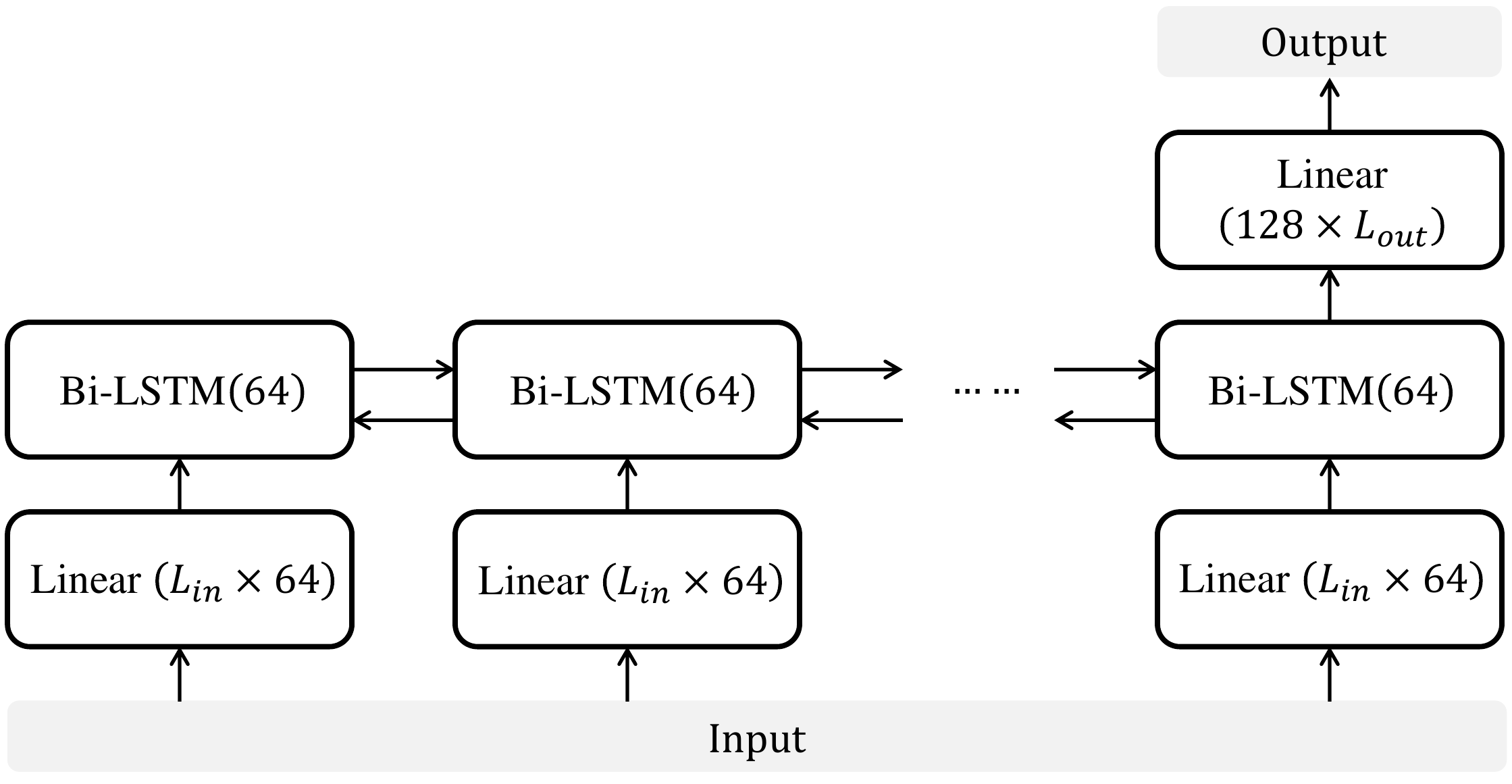}
\caption{The actor and critic networks are designed to be RNNs with Bi-LSTM units.}
\label{fig:4.4}
\end{figure}

\begin{rem}[Alternative observation design]
In prior works of AutoCA \cite{yu1,CR3} (without hidden terminals), the observation of a terminal is often designed as $\{a_n^t,\allowbreak~\text{sensed}\allowbreak~\text{channel}\allowbreak~\text{states},\allowbreak~\text{AP's}\allowbreak~\text{feedback}\}$. When applied in our problem, the observation would have three rows:
the first row consists of the actions of the $n$-th terminal itself,
the second row consists of the observed actions of the OH neighbors $o_{\text{OH}(n)}^\tau$,
and the third row consists of the feedback from the AP in the look-back window. Specifically, we would have
\begin{equation}\label{eq:obs1}
\widetilde{\mathcal{O}}_n^t=\{a_n^\tau,~o_{\text{OH}(n)}^\tau,~\text{ACK}^{\tau},:\tau=t-1,t-2,\ldots,t-W\}.
\end{equation}
The comparisons between \eqref{eq:obs} and \eqref{eq:obs1} are given in Section \ref{sec:V}.
\end{rem}

\subsection{Reward design}
In MADRL-HT, all agents learn simultaneously and their policies vary after each training. From a terminal's perspective, it faces a highly non-stationary environment, especially when there are hidden terminals whose dynamics are partially observable. To learn a good set of transmission policies, the reward design is of crucial importance.

Recall from \eqref{goalpi} that our goal is to maximize the $\alpha$-fairness function. This corresponds to an episodic MADRL task. That is, the terminals interact with each other in an episodic manner and an episode lasts for a fixed number of slots. A reward is generated at the end of each episode, which is the $\alpha$-fairness measure computed from \eqref{afairness}. The goal of MADRL is then to learn a set of policies for the terminals to maximize the reward generated in an episode. This formulation, however, is sample inefficient due to the very sparse reward -- we empirically found that it is difficult for the agents to learn good transmission policies to maximize the $\alpha$-fairness.

To mitigate this problem, we resort to a continuous MADRL formulation. Specifically, the terminals still interact with each other in an episodic manner and an episode lasts for a fixed number of slots. However, unlike the episodic formulation, we assign a global instantaneous reward $r_t$ to all the terminals at each slot $t$ during the episode, as opposed to generating a single reward at the end of the episode.
The instantaneous reward $r_t$ is specified below.

Let us define a flag $\delta_n^\tau\in\{0,1\}$ to indicate whether a packet (of length $D$) is successfully transmitted in $[\tau,\tau+D-1]$ from the $n$-th terminal. At the beginning of a slot $t$, the number of packets successfully transmitted by the $n$-th terminal during the look-back window is given by
\begin{equation}
    M_n^t=\sum_{\tau=t-W}^{t-1}\delta_n^\tau.
\end{equation}
Notice that $M_n^t$, $n=1,2,3,...,N$, is known to the AP, hence, can be used to generate the reward. Let $\mathcal{M}\triangleq\{M_n^t:n=1,2,\ldots,N\}$, we further define
\begin{equation*} G^t\triangleq\max\mathcal{M}-\min\mathcal{M},
\end{equation*}
where $G^t$ is the range of the set $\mathcal{M}$, i.e., the difference between the largest and smallest values of $\mathcal{M}$.

Then, the terminals make transmission decisions as long as they sense that the channel has been idle for at least a DIFS.
The global reward incurred by their actions in slot $t$ is generated at slot $t+D$ when $\delta^t_n$ is known. Let
\begin{equation*}
F^t\triangleq\delta_{1}^t\vee\delta_{2}^t\vee ... \vee\delta_{N}^t,
\end{equation*}
where $F^t$ indicates whether an ACK is received in slot $t+D$ ($F^t=1$) or not ($F^t=0$).

\begin{defi}[Reward]
\label{defirew}
The instantaneous global reward for slot $t$ is computed by
\begin{equation}\label{eq:reward}
    r^t\triangleq\begin{cases}
    +1, & \text{if $F^t=1$ and $G^t\leq 1$};\\
    +1, & \text{if $F^t=1$, $G^t>1$, and $\delta_{\arg\min\mathcal{M}}^t=1$};\\
    -1, & \text{if $F^t=1$, $G^t>1$, and $\delta_{\arg\min\mathcal{M}}^t=0$};\\
    -1, & \text{if $F^t=0$};\\
    0, & \text{if no terminal tranmits in slot $t$}.\\
    \end{cases}
\end{equation}
In particular, 
$\delta_{\arg\min\mathcal{M}}^t$ indicates whether the ($\arg\min\mathcal{M}$)-th terminal (i.e., the terminal with the least number of successful transmissions in the look-back window) successfully transmits a packet in slots $[t,t+D-1]$.
\end{defi}

The motivations behind \eqref{eq:reward} are as follows. First, our reward punishes actions $\{a_n^t:n=1,2,\ldots,N\}$ that lead to an unsuccessful transmission in slot $t$, i.e., the fourth line of \eqref{eq:reward} when $F^t=0$. Second, for a successful transmission $F^t=1$,
\begin{itemize}
    \item If $G^t\leq 1$, we consider that the current policies have achieved a certain degree of fairness, and hence, a positive reward is assigned to all terminals.
    \item If $G^t>1$, we consider that the current policies are unfair. In this case, if the transmitting terminal is the terminal with the least number of successful transmissions in the look-back window (i.e., $\delta_{\arg\min\mathcal{M}}^t=1$), we reward the  transmitting terminal as well as other terminals (for being idle) -- a positive reward is assigned to all terminals. In contrast, if $\delta_{\arg\min\mathcal{M}}^t=0$, we punish all terminals.
\end{itemize}
Finally, a zero reward is assigned to all terminals if no one transmits in slot $t$.

Overall, to attain a positive reward, the terminals must cooperate to 1) achieve an ACK instead of a NACK -- they must learn to maximize the throughput while avoiding collisions; 2) balance the transmission opportunities to achieve an almost equal number of successful transmissions in the look-back window -- intuitively, a terminal with less successful transmissions in the look-back window should learn to be more aggressive, while a terminal with more successful transmissions in the look-back window should learn to be less aggressive. 

With the above design principles, our reward function is aligned with the optimization objective of \eqref{goalpi}, and hence, can potentially learn a good set of transmission policies.

\begin{rem}[The $\alpha$-reward]
In addition to the reward given in Definition \ref{defirew}, we have considered many other forms of reward. In general, the reward can be global (where all terminals receive the same reward) or individual (where the terminals receive distinct rewards). We empirically found that global reward is a better fit for our design. For benchmark purposes, we introduce another global reward design, dubbed the $\alpha$-reward, that utilizes the $\alpha$-fairness of all terminals during the look-back window as the reward. Specifically, we define
\begin{equation}\label{eq:reward1}
    \widetilde{r}_t=\sum_{n=1}^N f(\mathcal{T}_n^t),
\end{equation}
where $\mathcal{T}_n^t$ is the throughput of the $n$-th terminal computed over the duration $[t-W,t-1]$, i.e., the look-back window of slot $t$. As can be seen, the $\alpha$-reward matches the objective \eqref{goalpi} straightforwardly -- a terminal is rewarded for the increase of throughput $\mathcal{T}_n^t$, but the reward gets smaller and smaller with the increase of $\mathcal{T}_n^t$ since $f$ is a concave function. The performances of \eqref{eq:reward} and \eqref{eq:reward1} will be compared in Section \ref{sec:V}.
\end{rem}

\subsection{MADRL-HT}
Given the definitions of partial observations and reward, this subsection presents the overall architecture of MADRL-HT for the AutoCA problem. Let us set out to define the actor and critic networks.

\begin{defi}[Actor]
The action of a terminal is determined by the transmission policy $\pi_n$. We parameterize this policy by a DNN, dubbed “actor”, with a vector of parameters $\bm{\theta_n}$. The actor takes the observation $o_n^t$ as input and outputs action $a_n^t$ for slot $t$, that is,
\begin{equation}
    a_n^t=\Psi_{\bm{\theta_n}}(\mathcal{O}_n^t).
\end{equation}
Each terminal is equipped with an actor.
\end{defi}

\begin{defi}[Critic]
We define a critic network to evaluate the value function of a global state $V(S^t)$. The critic network is parameterized by $\bm{\phi}$, giving
\begin{equation}
    V(S^t)=\Psi_{\bm{\phi}}(S^t).
\end{equation}
The critic is deployed at the AP.
\end{defi}

\begin{figure}[t]
\centering
\includegraphics[width=0.95\linewidth]{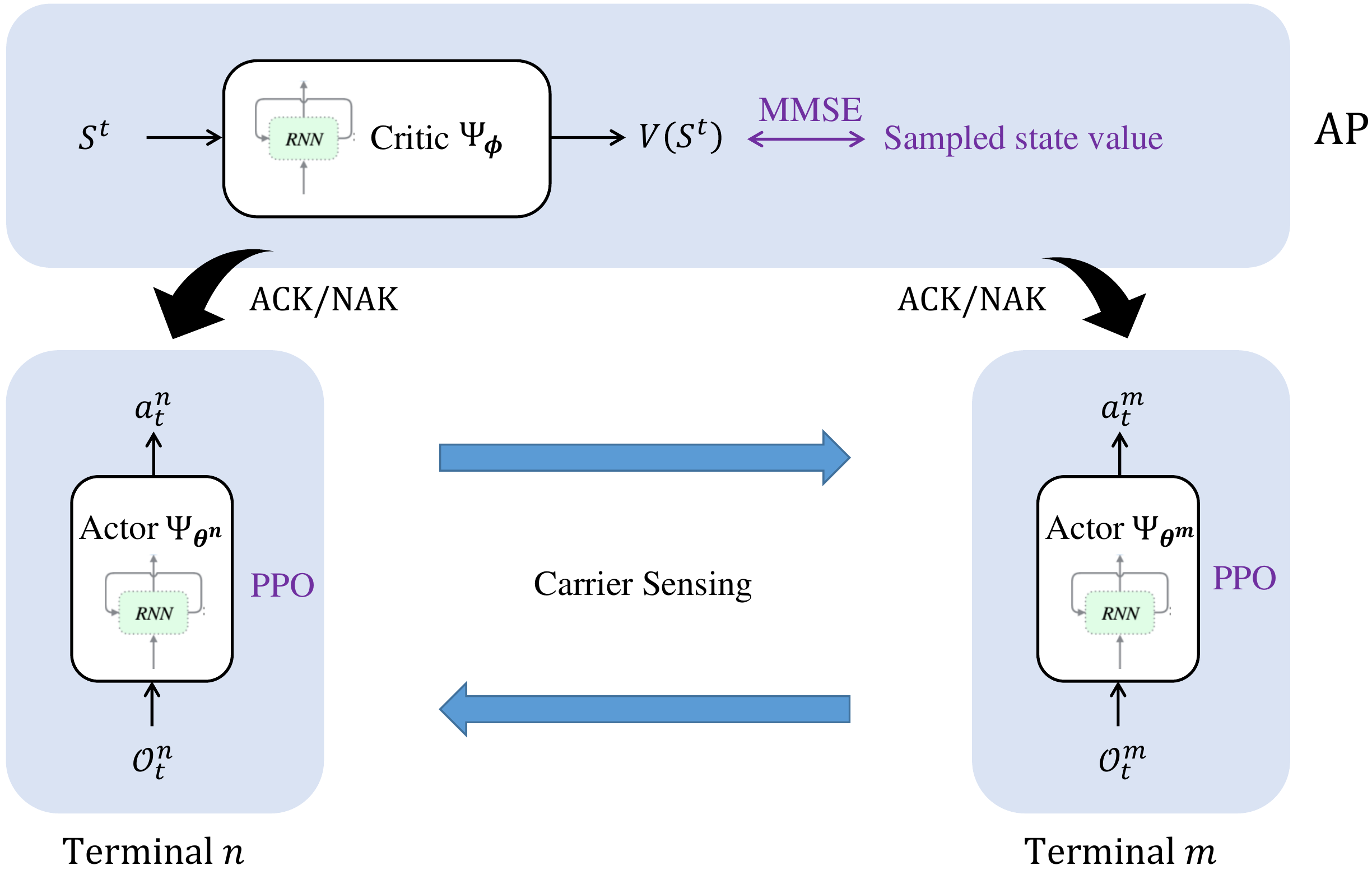}
\caption{The overall learning paradigm of MADRL-HT: terminals are equipped with actor networks and interact with each other in an episodic manner. The AP is equipped with a critic to estimate the state value function. At the end of an episode, the actors are trained by PPO and the critic is trained by regression on the MSE.}
\label{fig:4.5}
\end{figure}

Since the transmission history of each terminal is a temporal sequence, we design both actor and critic to be recurrent neural networks (RNNs) with bi-directional LSTM \cite{LSTM} as the recurrent units. As shown in Fig.~\ref{fig:4.4}, the actor and critic share the same architecture with the only difference being the input and output.
\begin{enumerate}
    \item {\it Input}. For the actor at the $n$-th terminal, the input is a $3\times W$ dimensional matrix consisting of the transmission history of the $n$-th terminal, $\text{OH}(n)$, and $\text{TH}(n)$ in the look-back window, as given in \eqref{eq:obs1}. For the critic, the input is a $N\times W$ dimensional matrix consisting of the transmission history of all the terminals in the look-back window, i.e., $S^t$ in \eqref{eq:state}.
    \item {\it DNN}. The first layer is a linear layer of $64$ neurons, where $L_{in}=3$ for actors and $L_{in}=N$ for the critic. The second layer is a bi-directional LSTM layer with 64 hidden neurons. The third layer is a linear layer of $128$ neurons, where $L_{out}=2$ for actors and $L_{out}=1$ for the critic. The ReLu activation function is applied to the output of each linear layer.
    \item {\it Output}. For the actor at the $n$-th terminal, the output is a vector indicating the probability of transmission or being idle, i.e., $\text{output}=[\Pr(a_n^t=0),\Pr(a_n^t=1)]^\top$.
    For the critic, the output is the estimated value function $V(S^t)$.
\end{enumerate}

\begin{algorithm}[t]
\caption{MADRL-HT for autonomous channel access in the presence of hidden terminals.}
\label{algo:MADRL}
\begin{algorithmic}[1]
\State{\underline{\textbf{Initialization}}:}
\State{Initialize $N$ terminals and their topology;}
\State{Initialize the length of the look-back window $W$;}
\State{Initialize the length of an episode $T_e$ and the number of episodes $K$;}
\State{The terminals randomly initialize the actor networks with weights $\{\bm{\theta_n^0}:n=1,2,\ldots,N\}$;}
\State{The AP randomly initialize the critic network with weights $\bm{\phi^0}$.}
\For{$k=0:K-1$}
    \State{\underline{\textbf{Interactions}}:}
    \parState{The terminals interact with each other for one episode of $T_e$ slots, with the current actor networks $\Psi_{\theta_n^0}$.}
    \For{$t=W:T_e-1$}
        \parState{The terminals update their observations $\{o_n^t\}$ from their carrier sensed information and the feedback from the AP, following Definition \eqref{eq:obs}.}
        \parState{The terminals perform the actions $\{a_n^t\}$ and record the action chosen probabilities $\Psi_{\bm{\theta_n^k}}(a_n^t\mid o_n^t),\forall n$.}
        \State{The AP computes the reward.}
    \EndFor
    \parState{At the end of the episode, the AP computes the advantage function $A_t,\forall t$ following \eqref{eq:adv} and broadcasts them to all terminals.}
    \State{\underline{\textbf{DNN training}}:}
    \State{Upon receiving $\{A_t\}$, each terminal updates its actor following \eqref{eq:ppo_ac};}
    \State{The AP updates the critic by the regression on MSE, as shown in \eqref{eq:ppo_vf}.}
\EndFor
\end{algorithmic}
\end{algorithm}

The overall learning paradigm is illustrated in Fig.~\ref{fig:4.5} and explained in detail in Algorithm~\ref{algo:MADRL}. As shown, the terminals execute their actor/policy networks to interact with each other in an episodic manner. During the episode, the terminals continuously update their observations based on their carrier sensed information and the feedback broadcasted by the AP. The AP, on the other hand, computes the reward according to \eqref{eq:reward}. Notice that a reward $r_t$ is computable only $D$ slots after the decision is made at the beginning of slot $t$.

At the end of an episode, the AP collects the actions of the terminals to form the global state $S^t$ and estimates the state-value function of $S^t$ from the current critic. Then, the AP computes the advantage function for each global state $S^t$, giving\footnote{In the implementation, we use generalized advantage estimation (GAE) \cite{GAE} to compute the advantage functions to make the training more stable.}
\begin{equation}\label{eq:adv}
A_t=\sum_{t'>t}\gamma^{t'-t}r^{t'}-\Psi_{\bm{\theta_k}}(S^t),\quad\forall t,
\end{equation}
where $\gamma$ is a discounting factor, and broadcasts them to all terminals.

The update of actor and critic networks follows the proximal policy optimization (PPO) algorithm \cite{PPO}. Specifically, upon receiving $\{A_t\}$, each terminal updates its actor by maximizing a PPO-clip objective:
\begin{equation}\label{eq:ppo_ac}
    \bm{\theta_n^{k+1}}=\arg\max_{\bm{\theta_n}}\sum_{t=W}^{T_e-1}\min\left(\frac{\Psi_{\bm{\theta_n}}(a_n^t\mid o_n^t)}{\Psi_{\bm{\theta_n^k}}(a_n^t\mid o_n^t)}A_t,g(\epsilon,A_t)\right),
\end{equation}
where
\begin{equation*}
    g(\epsilon,A)=\begin{cases}
        (1+\epsilon)A, & A\geq 0;\\
        (1-\epsilon)A, & A<0,
    \end{cases}
\end{equation*}
and $\epsilon$ is a clipping ratio.

On the other hand, the AP updates the critic by minimizing the MSE between the estimated state value $\Psi_{\bm{\phi}}(S^t)$ and the sampled state value from the interactions:
\begin{equation}\label{eq:ppo_vf}
    \bm{\phi^{k+1}}=\arg\min_\phi\sum_{t=W}^{T_e}\left(\sum_{t'>t}\gamma^{t'-t}r^{t'}-\Psi_\phi(S^t)\right).
\end{equation}
It is worth noting that our MADRL-HT solution can also operate in a centralized training and distributed execution manner, in which case the AP trains both the actors and critic and periodically transmits the latest actor networks to the terminals.

\begin{rem}
In our MADRL-HT approach, a sequence of $W$ slots of history is fed into the actor and decoder for decision making and value function estimation, respectively. In contrast, the widely-adopted learning architecture in the literature \cite{drqn,coma,qmix} assumes that the past history has been captured by the hidden state of the RNNs and feeds only the latest slot into the RNNs for decision making.
We point out that this widely-used architecture cannot be used in our problem because of the partial observability of each terminal.
In our solution, the observations and rewards of a terminal are updated based on the feedback from the AP (see Eqn.~\eqref{eq:reward} and Algorithm~\ref{algo:observations}), which is delayed for $D$ slots. As a result, we have to revisit and modify past observations and rewards upon receiving the ACK/NACK feedback.
\end{rem}

\begin{table}
\renewcommand*{\arraystretch}{0.9}
    \caption{A list of hyper-parameters.}
    \label{tab:1}
    \centering
    \begin{tabular}{lll}
        \toprule
        \multicolumn{2}{c}{Hyperparameters} & Values                                                       \\
        \midrule
        \multirow{3}{*}{System}             & Length of a slot                                 & 9$\mu s$  \\
                                            & Packet length $D$                                & 5 slots   \\
                                            & Look-back window $W$                             & 40 slots  \\
        \cmidrule{1-3}
        \multirow{7}{*}{PPO}                & Number of epochs $K$                             & 10000     \\
                                            & Number of slots/episode $T_e$                & 100 slots \\
                                            & Clipping ratio $\varepsilon$                     & 0.2       \\
                                            & Learning rate for actor                          & 0.001     \\
                                            & Learning rate for critic                         & 0.0005    \\
                                            & Discount factor $\gamma$                      & 0.99      \\
                                            & Lambda for GAE $\lambda$                  & 0.95      \\
        \cmidrule{1-3}
        \multirow{3}{*}{CSMA/CA}               & Minimum CW: $\text{CW}_0$          & 2 slots   \\
                                            & Maximum CW: $\text{CW}_\text{max}$ & 128 slots \\
                                            & DIFS                                             & 1 slot    \\
        \bottomrule
    \end{tabular}
\end{table}

\begin{table}[t]
    \caption{A summary of the topologies considered in the experiments.}
    \setlength{\tabcolsep}{5mm} 
    \label{tab:2}
    \centering
    \begin{tabular}{ll}
        \toprule
        Abbreviations & Topological Structures \\
        \midrule
        ~~\quad Topo2 & $\{A,B\}$  \\
        ~~\quad Topo2' & $\{A\mid B\}$ \\
        \midrule
        ~~\quad Topo3 & $\{A,B,C\}$ \\
        ~~\quad Topo3' & $\{A,B\mid C\}$ \\
        ~~\quad Topo3''  & $\{A,B\mid B,C\}$ \\
        \midrule
        ~~\quad Topo4   & $\{A,B,C,D\}$ \\
        ~~\quad Topo4'  & $\{A,B,C\mid D\}$ \\
        ~~\quad Topo4'' & $\{A,B\mid B,C\mid D\}$ \\
        \bottomrule
    \end{tabular}
\end{table}

\section{Numerical Experiments}\label{sec:V}
\begin{figure*}
    \centering
    \begin{subfigure}[t]{0.3\linewidth}
        \centering
        \includegraphics[width=\linewidth]{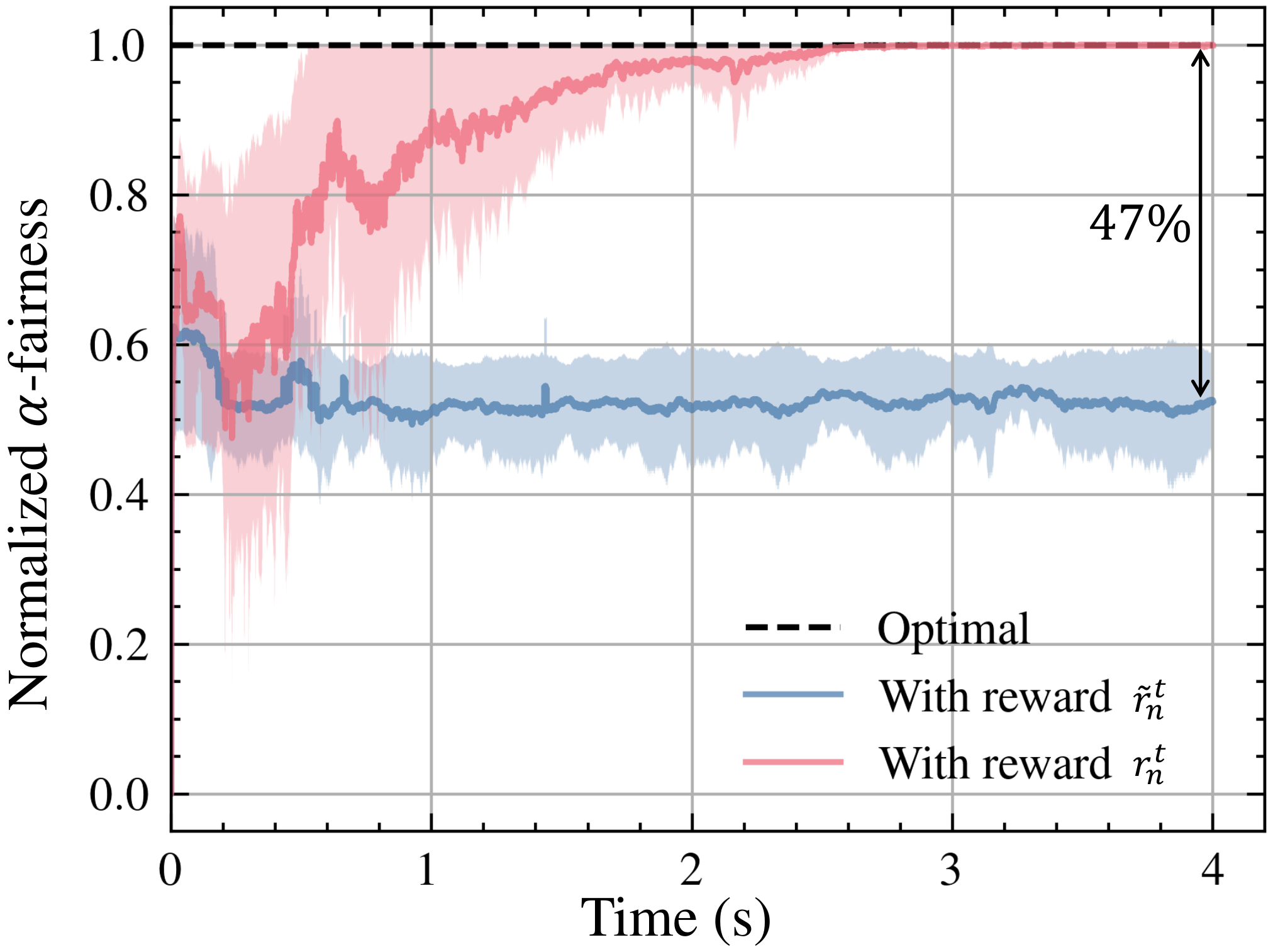}
        \caption{Normalized $\alpha$-fairness}
    \end{subfigure}
    \begin{subfigure}[t]{0.3\linewidth}
        \centering
        \includegraphics[width=\linewidth]{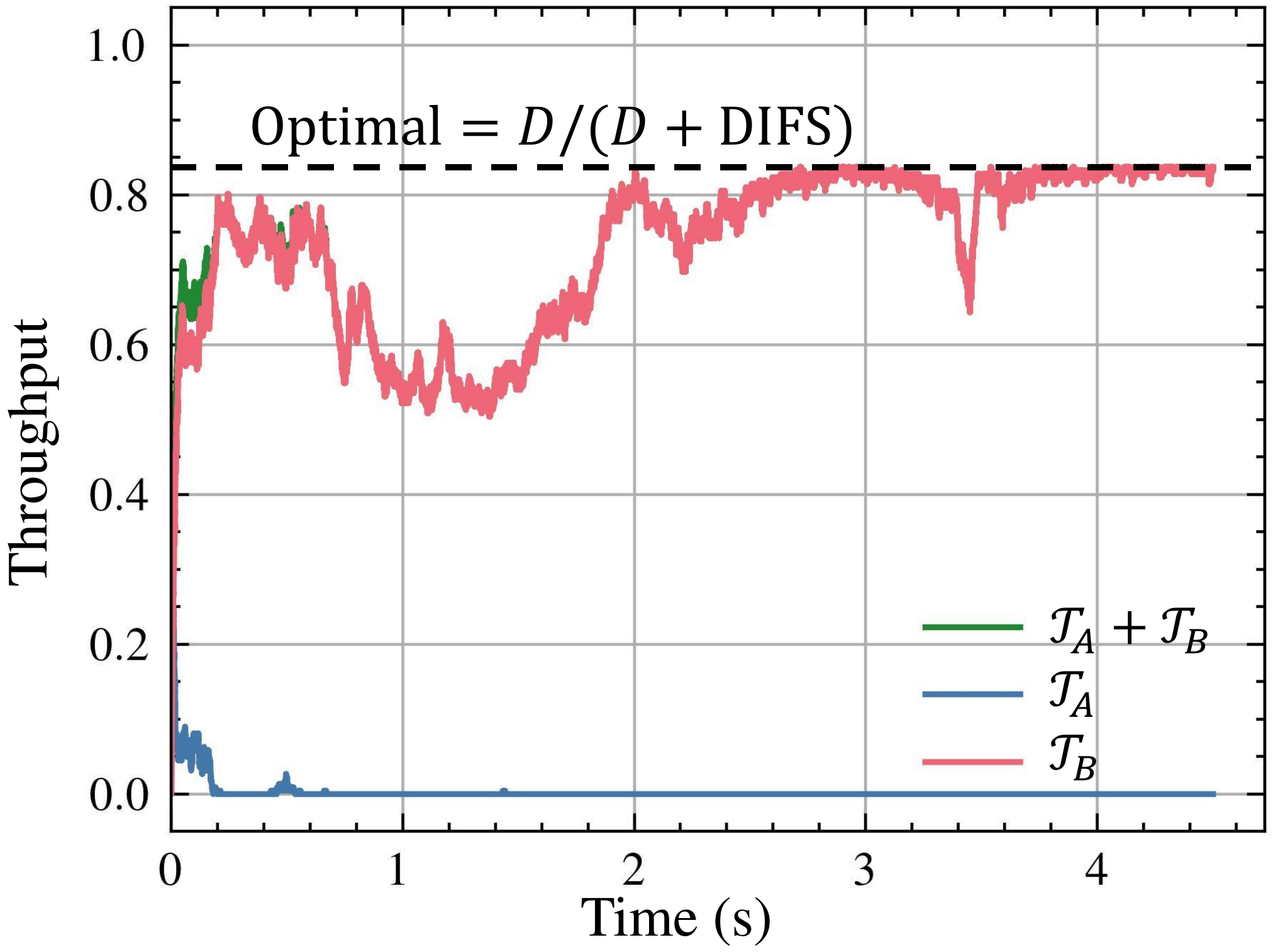}
        \caption{Throughput with $\widetilde{r}_t$}
    \end{subfigure}
    \begin{subfigure}[t]{0.3\linewidth}
        \centering
        \includegraphics[width=\linewidth]{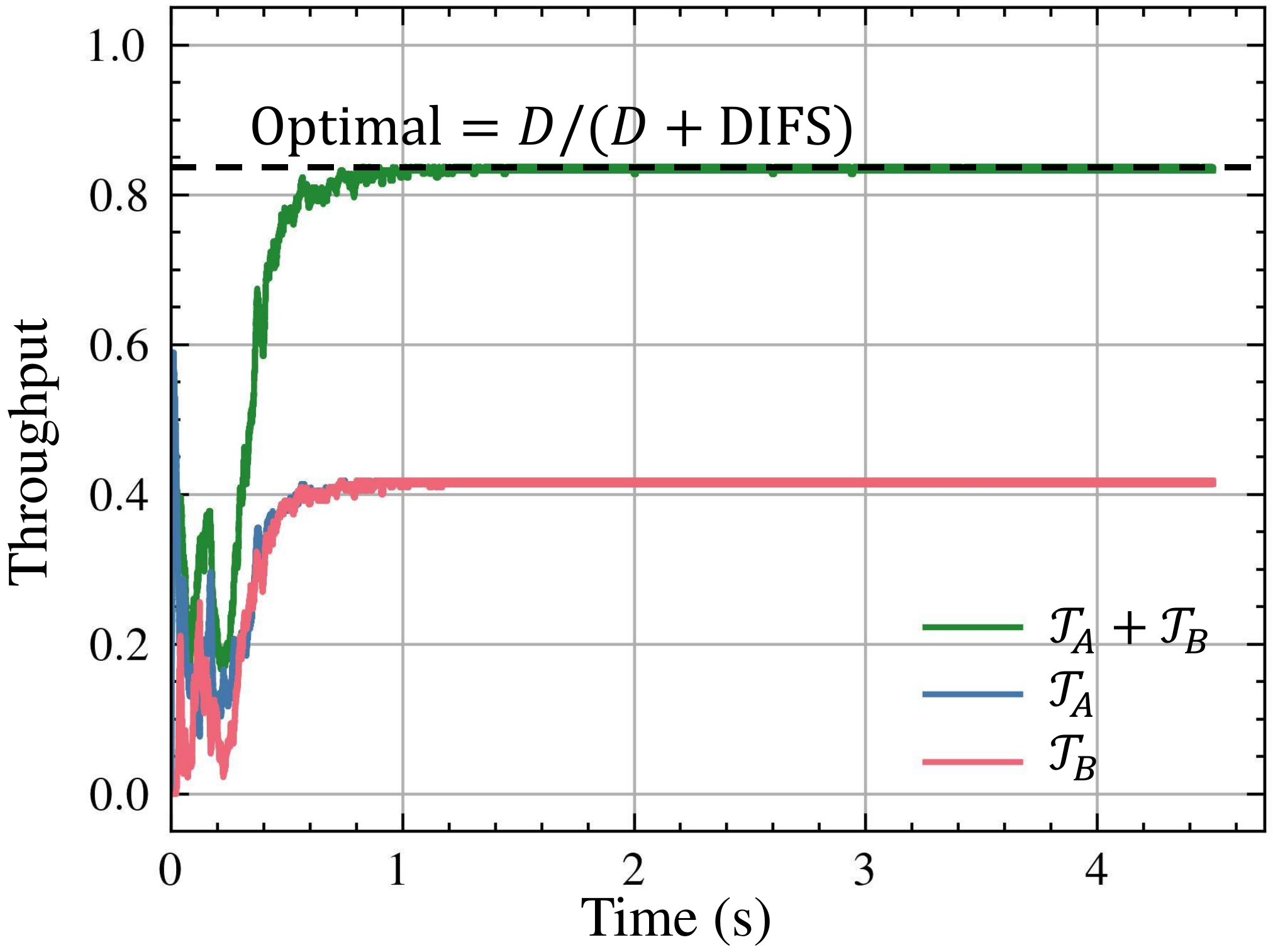}
        \caption{Throughput with $r_t$}
    \end{subfigure}
    \caption{Performance of MADRL-HT in Topo2 with the proposed reward design $r_t$ and the conventional reward design $\widetilde{r}_t$: (a) Normalized $\alpha$-fairness; (b) Throughput with $\widetilde{r}_t$; (c) Throughput with $r_t$.}
    \label{fig:S0}
\end{figure*}

This section performs extensive numerical experiments to verify the performance of our MADRL-HT solution. We consider two benchmarks: one is the optimal set of transmission policies derived for the topologies considered in the experiments; the other is the CSMA/CA protocol, with which a terminal senses the common channel before accessing it and retransmits -- in the case of a collision -- after an exponential backoff.

\subsection{Experimental setup}
The hyper-parameter settings for the experiments are shown in Table \ref{tab:1} unless specified otherwise. For the Wi-Fi BSS, we assume that one slot lasts for $9~\mu s$ and a packet lasts for $5$ slots. A terminal can be either an intelligent terminal operated with the actor of MADRL-HT or a standard terminal operated with CSMA/CA. The intelligent terminals make transmission decisions based on the transmission history of the last $W = 40$ slots. A standard terminal, on the other hand, follows the CSMA/CA protocol. In particular, we set the minimum and maximum contention windows of CSMA/CA to $\text{CW}_0=2$ slots and $\text{CW}_\text{max}=128$ slots, respectively. The DIFS is set to $1$ slot.

The topologies of the BSS considered in the experiments are summarized in Table \ref{tab:2}, in which uppercase letters stand for terminals in the BSS and we use vertical lines to separate TH neighbors. In Topo3', for example, $\{A,B\mid C\}$ means there are three terminals in the BSS: terminals A and B are OH neighbors to each other, while terminal C is a TH neighbor to both A and B.

\begin{figure*}
    \centering
    \begin{subfigure}[t]{0.3\linewidth}
        \centering
        \includegraphics[width=\linewidth]{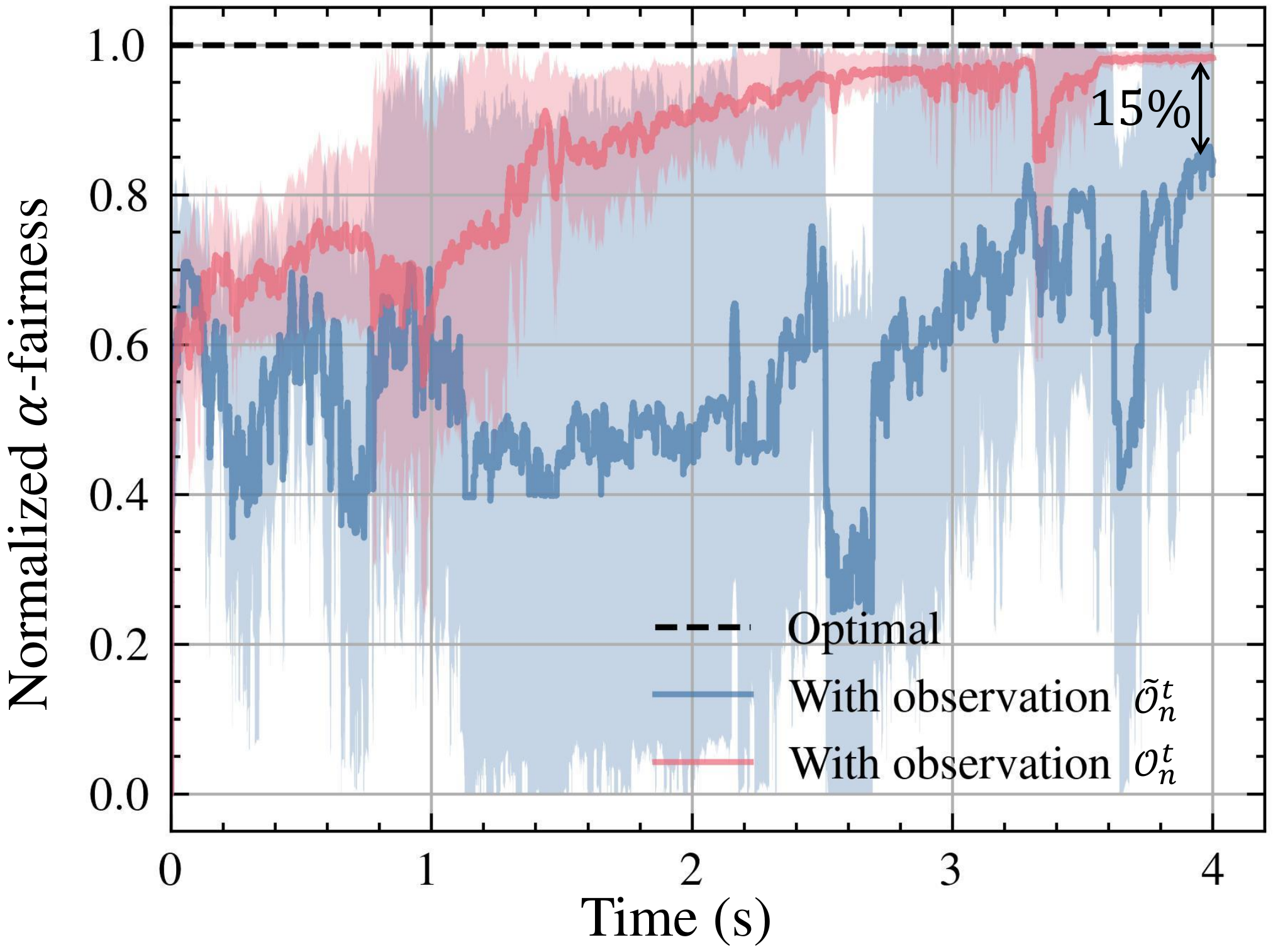}
        \caption{Normalized $\alpha$-fairness}
    \end{subfigure}
    \begin{subfigure}[t]{0.3\linewidth}
        \centering
        \includegraphics[width=\linewidth]{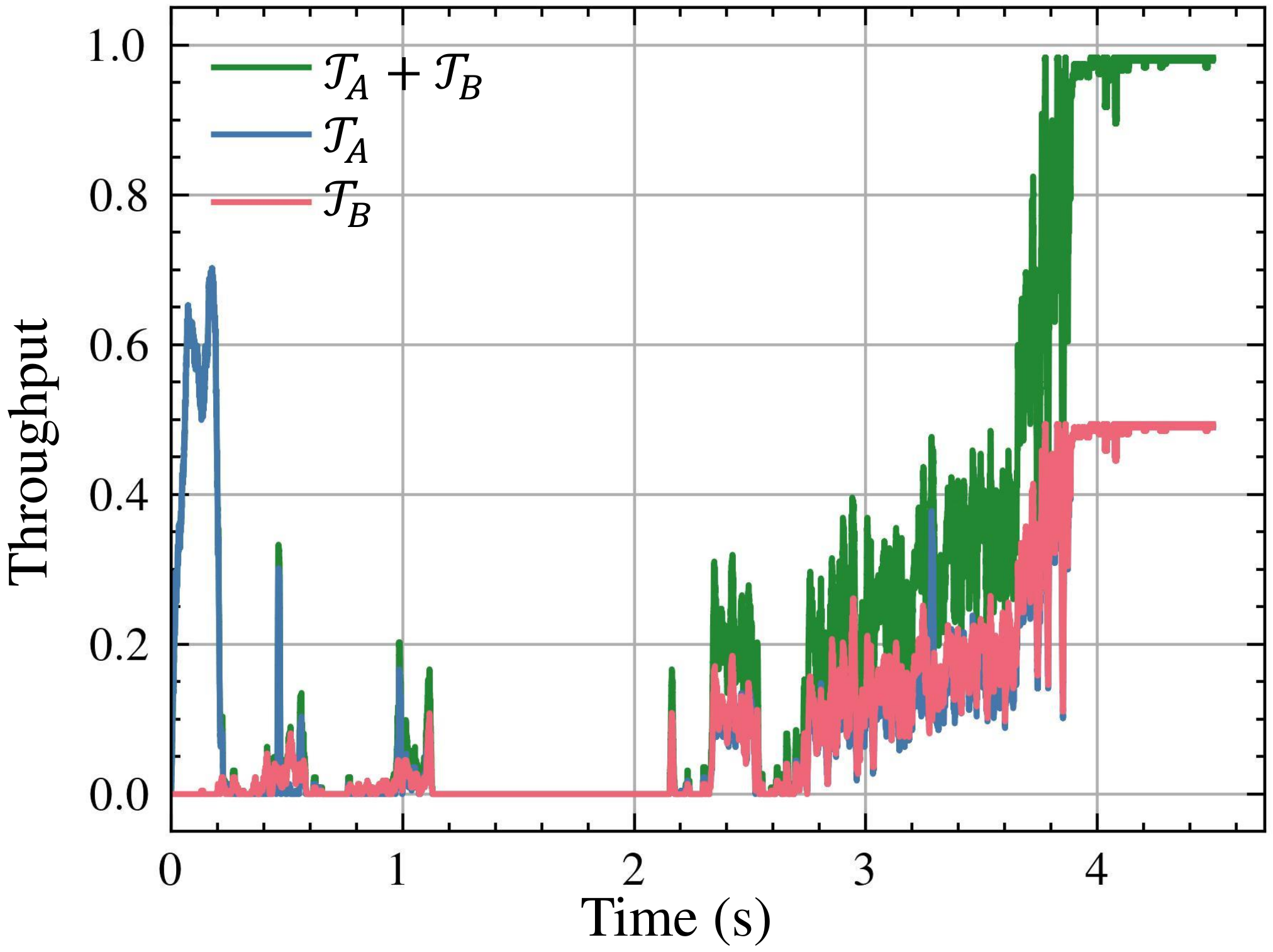}
        \caption{Throughput with $\widetilde{\mathcal{O}}_n^t$}
    \end{subfigure}
    \begin{subfigure}[t]{0.3\linewidth}
        \centering
        \includegraphics[width=\linewidth]{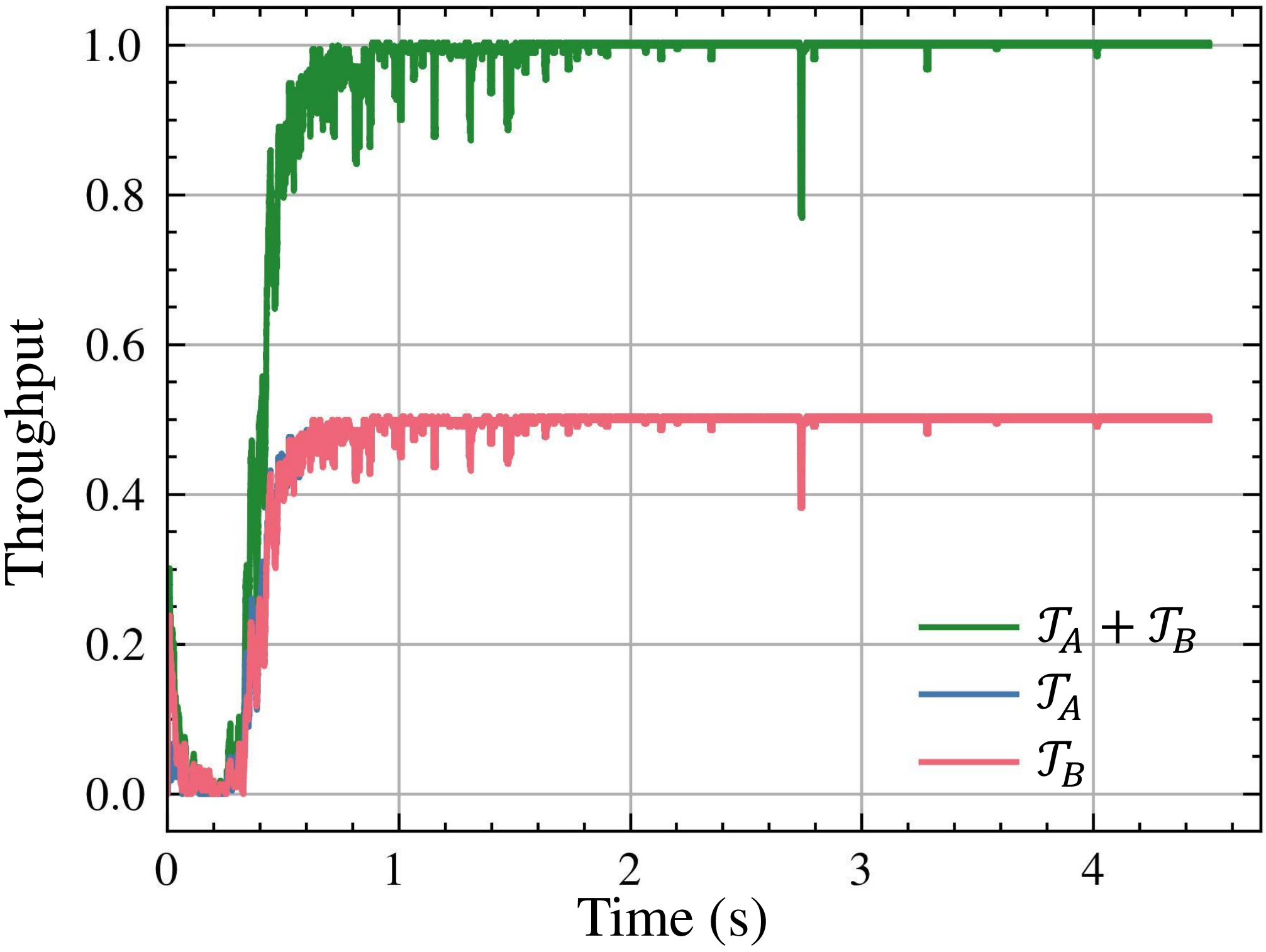}
        \caption{Throughput with $\mathcal{O}_n^t$}
    \end{subfigure}
    \caption{Performance of MADRL-HT in Topo2' with the proposed observation design $\mathcal{O}_n^t$ and the conventional observation design $\widetilde{\mathcal{O}}_n^t$: (a) Normalized $\alpha$-fairness; (b) Throughput with $\widetilde{\mathcal{O}}_n^t$; (c) Throughput with $\mathcal{O}_n^t$.}
    \label{fig:S4}
\end{figure*}

For a given set of transmission policies $\bm{\pi}$, the key performance indicator is the $\alpha$-fairness $F(\bm{\pi})$, where $\alpha$ is set to $1$ (proportional fairness). Therefore, we have
\begin{equation}
F(\bm{\pi})=\sum_{n=1}^{N}\log(\mathcal{T}_n(\bm{\pi})+\varepsilon),
\end{equation}
where $\varepsilon = 0.001$ is a constant to ensure that $F(\bm{\pi})$ does not go to $-\infty$.
In addition to $\alpha$-fairness, other metrics, such as throughput of individual terminals, packet collision rate, packet delay and jitter, are also considered to analyze the learned behaviors of the terminals.

\begin{rem}[Short-term fairness versus long-term fairness]
It is worth noting that $\alpha$-fairness is a function of the evaluation period $T$, as shown in \eqref{eq:throughput}. In general, the short-term fairness with a small $T$ is much more difficult to achieve than the long-term fairness with a large $T$.
For a fair comparison in the experiments, the evaluation duration $T$ is set to $0.01$s to compute the throughput $\mathcal{T}_n(\bm{\pi})$ and $\alpha$-fairness for different schemes.
\end{rem}

\subsection{Reward function and observation design}
We first perform experiments to validate our designs of reward and observation, respectively.

In the first experiment, we evaluate our reward function $r_t$ in \eqref{eq:reward} benchmarked against the $\alpha$-reward $\widetilde{r}_t$ in \eqref{eq:reward1}. The experiments are performed on the topology Topo2, i.e., there are two terminals A and B in the Wi-Fi BSS and they are OH neighbors to each other.

Experimental results are presented in Fig.~\ref{fig:S0}(a), where the $x$-axis is time (i.e., the number of slots consumed in training) and the $y$-axis is the normalized $\alpha$-fairness achieved by the learned actors at the corresponding time. In particular, the $\alpha$-fairness is normalized to the region $[0,1]$. The lower bound `0' corresponds to the performance of a set of never-transmit policies, under which the terminals will never transmit and $F_\text{LB}(\bm{\pi})=2\log \varepsilon$. The upper bound `1', on the other hand, corresponds to the performance of the optimal policy. In Topo2, an optimal transmission pattern is `A0B0A0B0A0B0...', where '0' stands for the DIFS. This is because terminals A and B are OH neighbors and are subject to the LBT constraint. Therefore, the optimal BSS throughput of Topo2 is $D/(D+\text{DIFS})\approx 5/6$ and the optimal $\alpha$-fairness measure is $F_\text{UB}(\bm{\pi})=2\log (\frac{5}{12}+\varepsilon)$.

For both reward designs $r_t$ and $\widetilde{r}_t$, we have run the experiments several times and presented the mean (the solid lines in Fig.~\ref{fig:S0}(a)) and standard deviation (the shaded areas in Fig.~\ref{fig:S0}(a)) of the achieved $\alpha$-fairness. 
As can be seen, MADRL-HT cannot learn a good set of transmission policies with the $\alpha$-reward $\widetilde{r}_t$, despite the more intuitive interpretation of the reward function. In comparison, the proposed reward function $r_t$ achieves the optimal $\alpha$-fairness -- the performance gain over the $\alpha$-reward is up to $47\%$ when the learning converges.

The throughput achieved by each terminal in Fig.~\ref{fig:S0}(a) is more revealing to understand why the $\alpha$ reward does not yield good $\alpha$-fairness.
In Fig.~\ref{fig:S0}(b-c), we choose two typical experiments in Fig.~\ref{fig:S0}(a) and study the throughput of each terminal over the course of training. 
In particular, Fig.~\ref{fig:S0}(b) adopts the $\alpha$-reward $\widetilde{r}_t$, and Fig.~\ref{fig:S0}(c)  adopts the proposed reward $r^t$.
As shown, the BSS (sum) throughput can be maximized with both reward designs. Nevertheless, the $\alpha$-reward leads to a set of policies that are quite unfair – terminal A monopolizes the channel all the time, while terminal B has no transmission opportunities. In comparison, our reward design yields fair transmission policies -- the terminals have equal access to the common channel and the BSS throughput is maximized.

\begin{figure*}
    \centering
    \begin{subfigure}[t]{0.3\linewidth}
        \centering
        \includegraphics[width=\linewidth]{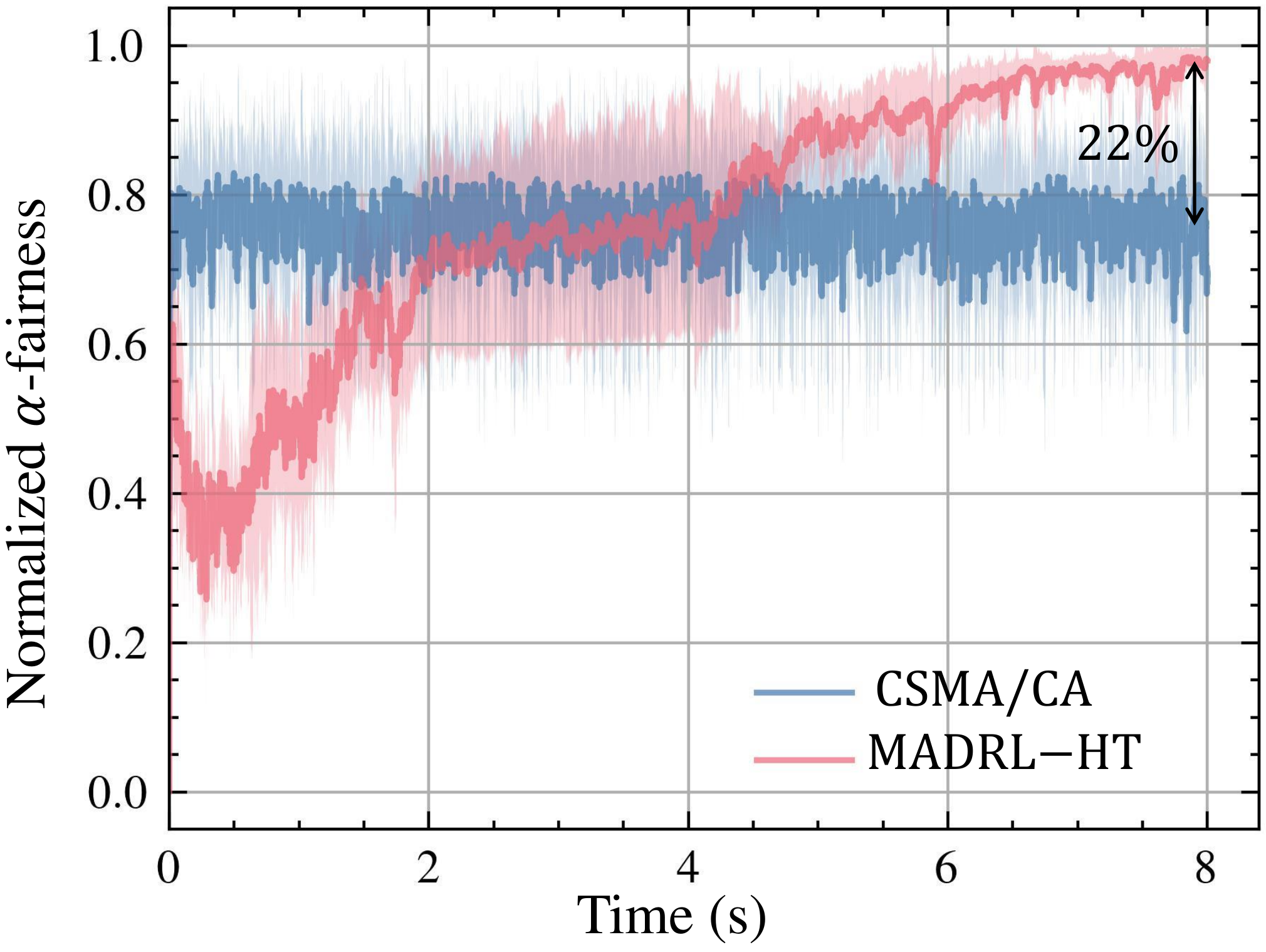}
        \caption{Topo3}
    \end{subfigure}
    \begin{subfigure}[t]{0.3\linewidth}
        \centering
        \includegraphics[width=\linewidth]{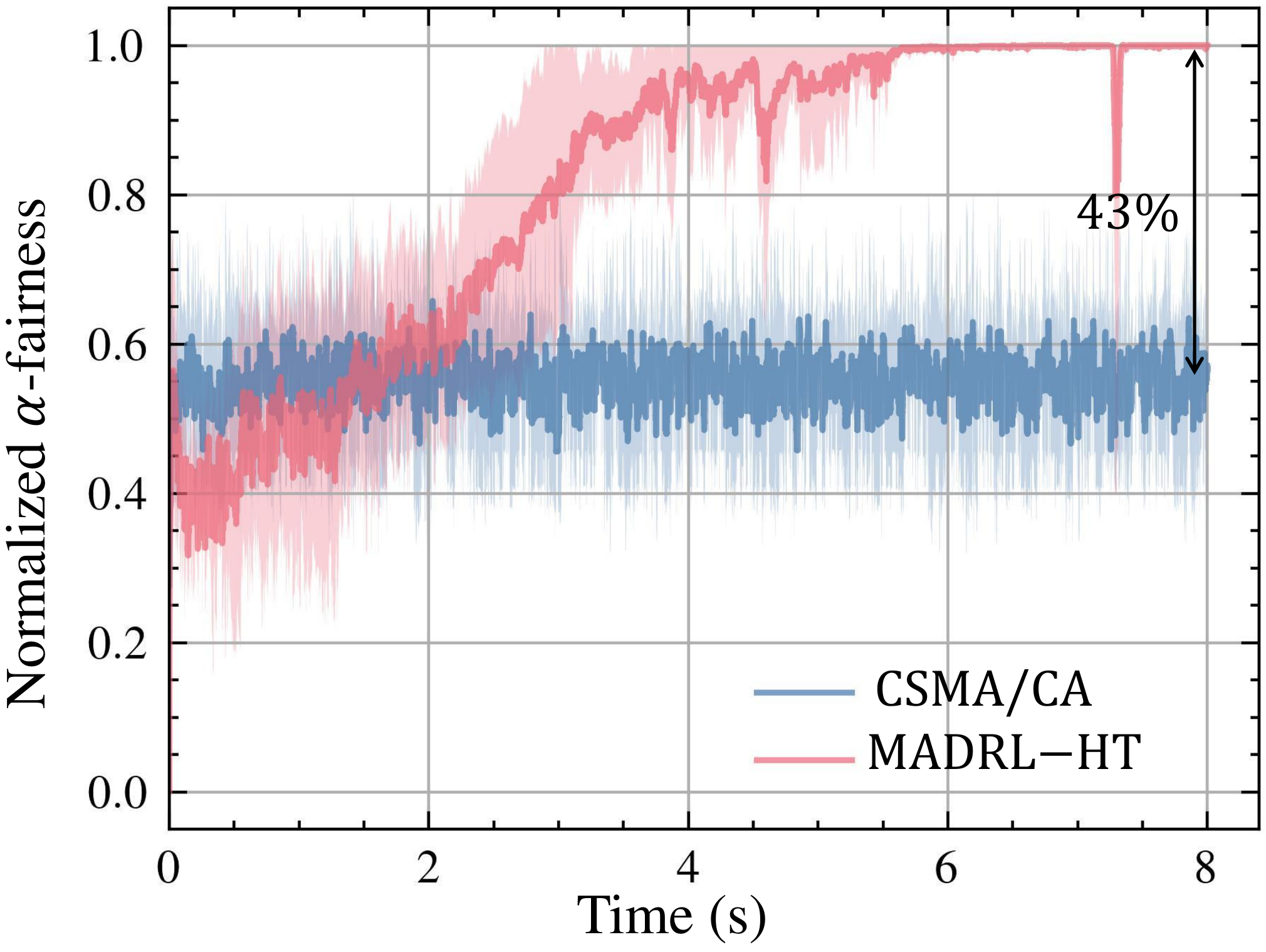}
        \caption{Topo3'}
    \end{subfigure}
    \begin{subfigure}[t]{0.3\linewidth}
        \centering
        \includegraphics[width=\linewidth]{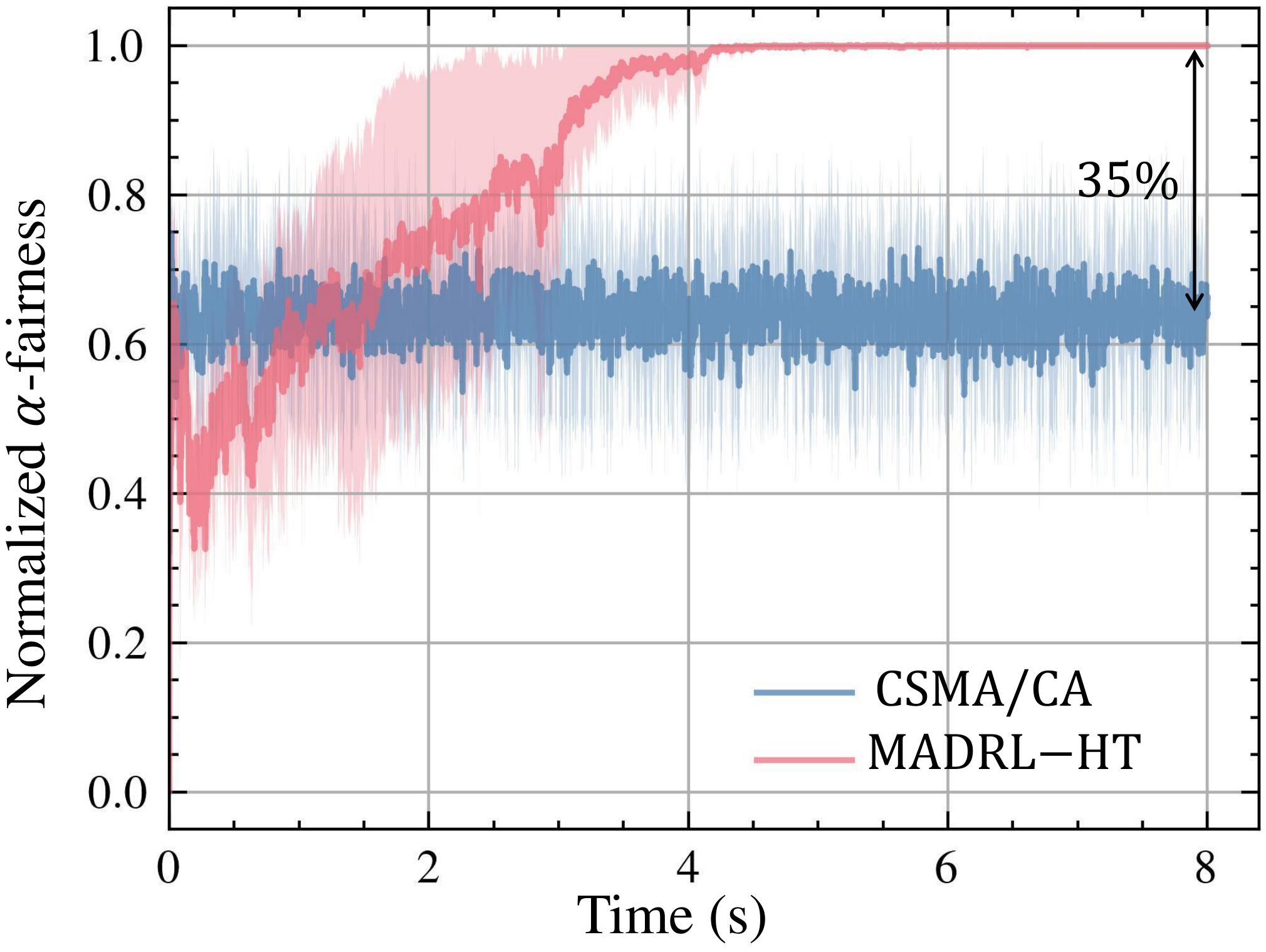}
        \caption{Topo3''}
    \end{subfigure}\\
    \begin{subfigure}[t]{0.3\linewidth}
        \centering
        \includegraphics[width=\linewidth]{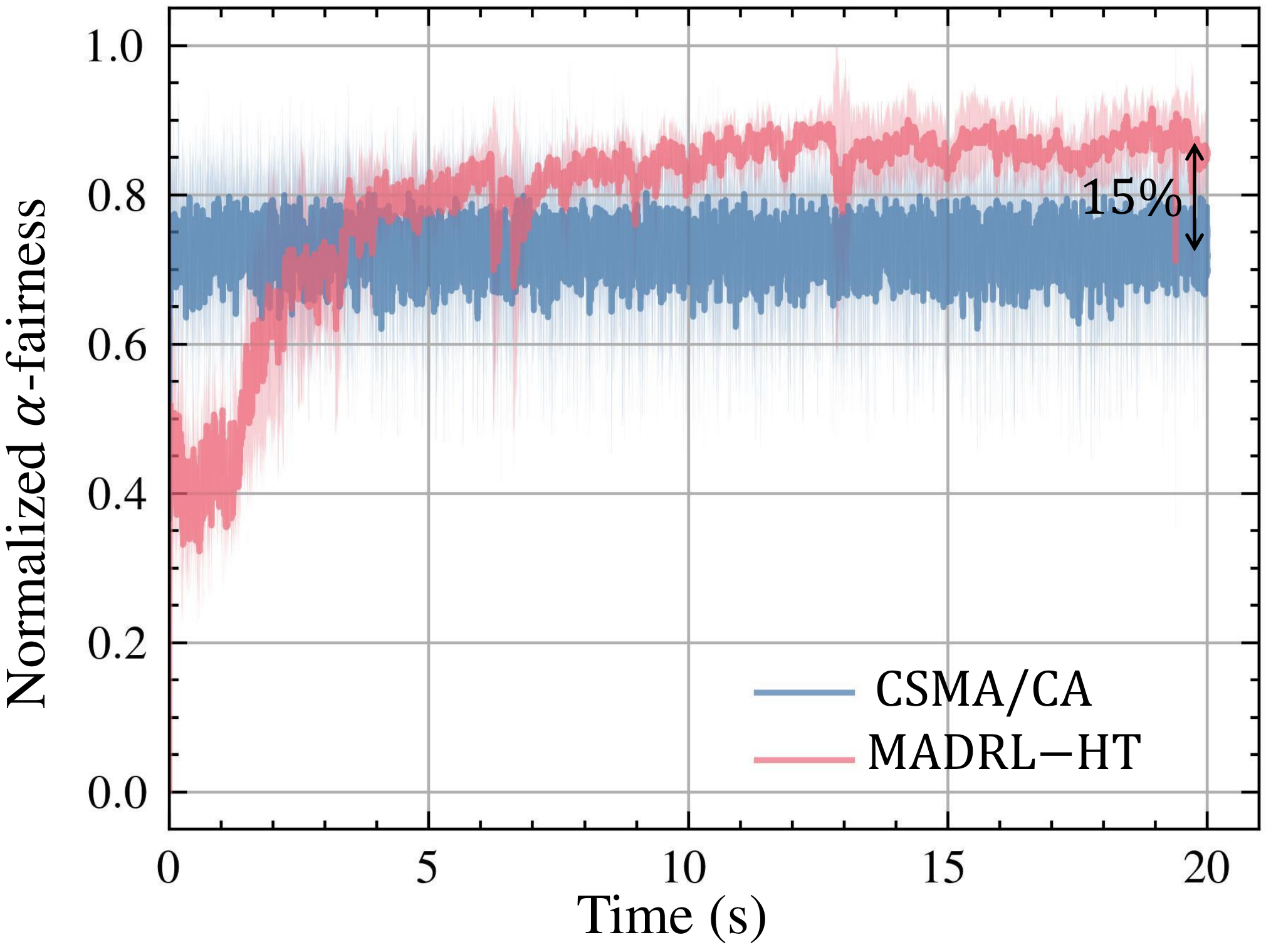}
        \caption{Topo4}
    \end{subfigure}
    \begin{subfigure}[t]{0.3\linewidth}
        \centering
        \includegraphics[width=\linewidth]{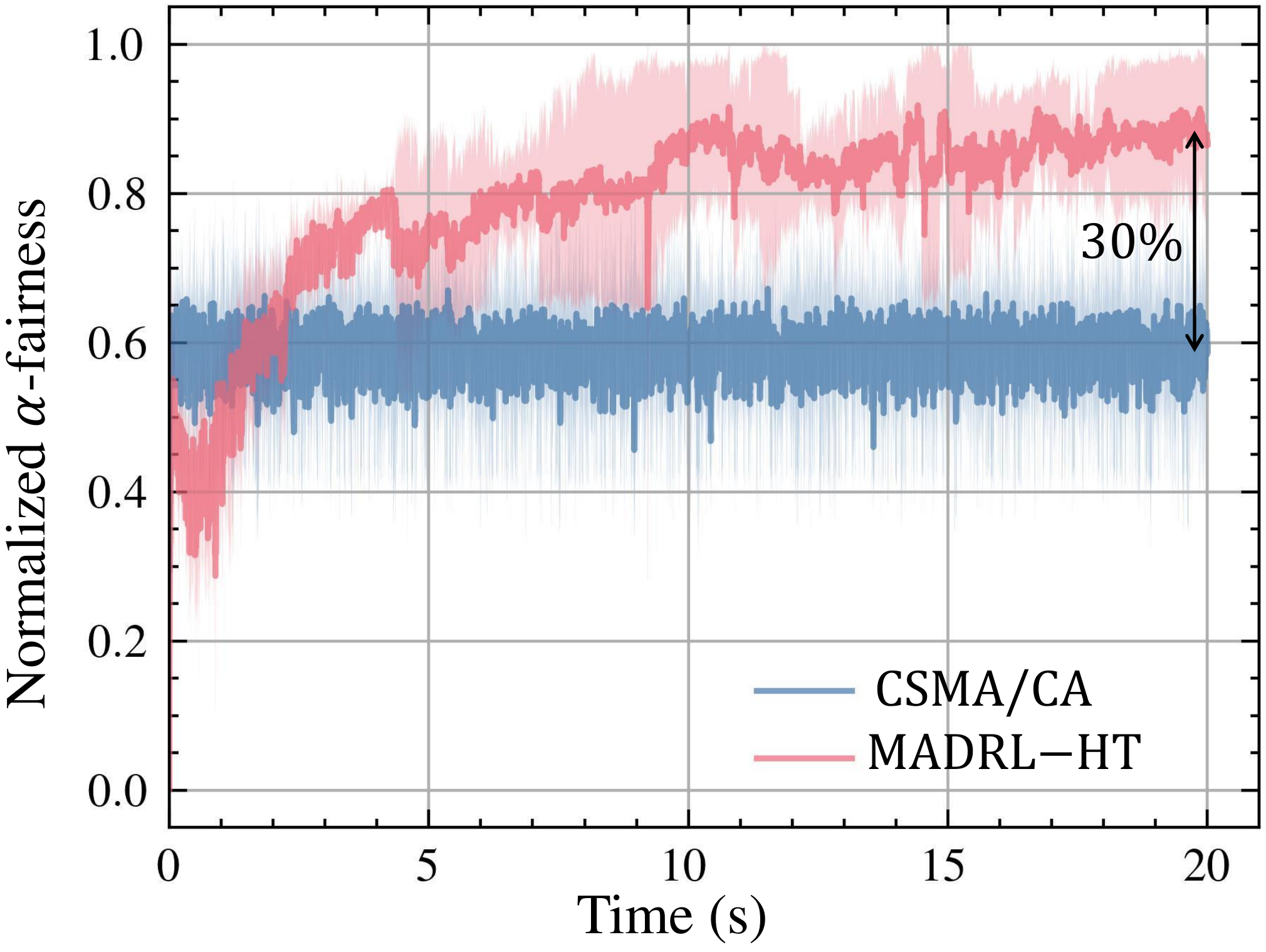}
        \caption{Topo4'}
    \end{subfigure}
    \begin{subfigure}[t]{0.3\linewidth}
        \centering
        \includegraphics[width=\linewidth]{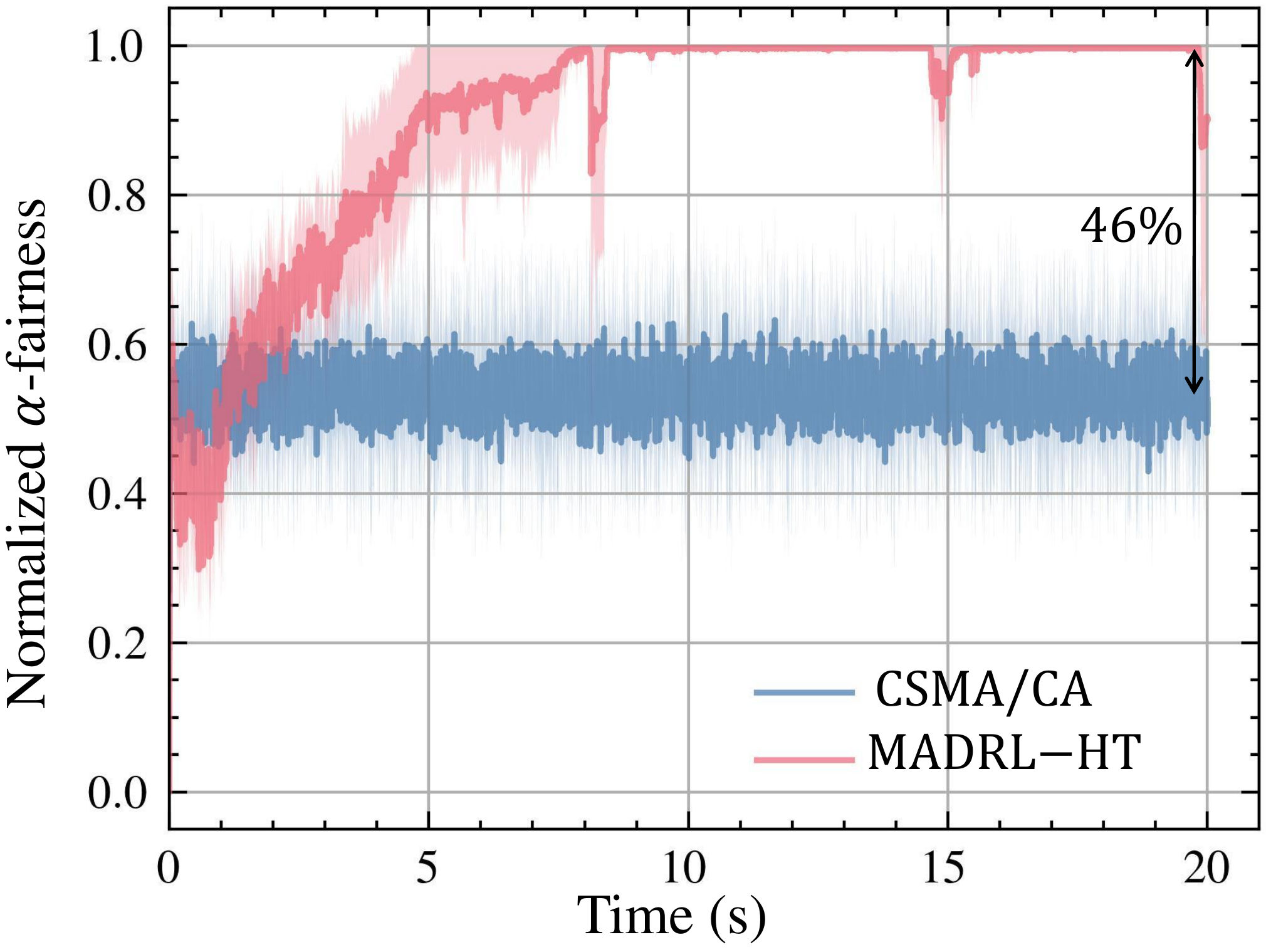}
        \caption{Topo4''}
    \end{subfigure}
    \caption{The normalized $\alpha$-fairness performance of a homogeneous MADRL-HT BSS and a standard CSMA/CA BSS in various topologies.}
    \label{fig:S6}
\end{figure*}

In the second experiment, we evaluate our observation design $\mathcal{O}_n^t$ in \eqref{eq:obs} benchmarked against the conventional design $\widetilde{\mathcal{O}}_n^t$ in \eqref{eq:obs1}. The experiments are performed on the topology Topo2', i.e., there are two terminals A and B in the Wi-Fi BSS and they are TH neighbors to each other.

The normalized $\alpha$-fairness versus time performance is shown in Fig.~\ref{fig:S4}(a), where the $\alpha$-fairness is normalized to $[0,1]$ as Fig.~\ref{fig:S0}(a). 
Unlike Fig.~\ref{fig:S0}(a), the maximum achievable $\alpha$-fairness in Topo2', i.e., what the normalized '1' stands for, is $F_\text{UB}(\bm{\pi})=2\log (\frac{1}{2}+\varepsilon)$.
The reason is that an optimal transmission pattern in Topo2' is `ABABAB...', since terminals A and B are TH neighbors -- the sensed channel is always idle for each of them and it is possible for them to transmit alternately with no DIFS. Therefore, the optimal BSS throughput is $1$ and the optimal $\alpha$-fairness is $2\log (\frac{1}{2}+\varepsilon)$.

As can be seen from Fig.~\ref{fig:S4}(a), with the proposed observation design, MADRL-HT achieves much faster and more reliable learning. The average $\alpha$-fairness performance of our design is $15\%$ better than the conventional design when the learning converges.
In addition to $\alpha$-fairness, we choose two specific experiments and study the throughput of each terminal in Fig.~\ref{fig:S4}(b-c).
In the chosen experiments, MADRL-HT can learn close-to-optimal policies with both observation designs, but our design yields much faster learning. We have further observed that, in more complex topologies  that involve more terminals, the conventional observation design can lead to very poor performance.

\subsection{MADRL-HT versus CSMA/CA}
In this section, we perform extensive experiments under various topologies to evaluate the performance of our MADRL-HT solution to the AutoCA problem.
Specifically, we consider a MADRL-HT BSS, where all terminals are intelligent, benchmarked against a standard CSMA/CA BSS, where all terminals are operated with the CSMA/CA protocol.

{\it 1) Normalized $\alpha$-fairness.}
Fig.~\ref{fig:S6} presents the normalized $\alpha$-fairness performance of both BSSs under various topologies listed in Table \ref{tab:2}. As Fig.~\ref{fig:S0}, we run each experiment multiple times and present the mean and standard deviation performances.
With our MADRL-HT solution, the terminals in the BSS start from random channel-access policies and arrive at a set of policies that is remarkably better than the CSMA/CA protocol in various topologies.
In terms of the average $\alpha$-fairness performance, the gains of the MADRL-HT BSS over the CSMA/CA BSS are up to $46\%$.

To better understand the learned behaviors of the MADRL-HT terminals, next we focus on two specific topologies: Topo4 and Topo4', and analyze other quality-of-service (QoS) metrics, i.e., the achieved throughput of each terminal, packet collision rate, average packet delay and jitter, for both BSSs. 

\begin{figure*}
    \centering
    \begin{subfigure}[t]{0.245\linewidth}
        \centering
        \includegraphics[width=\linewidth]{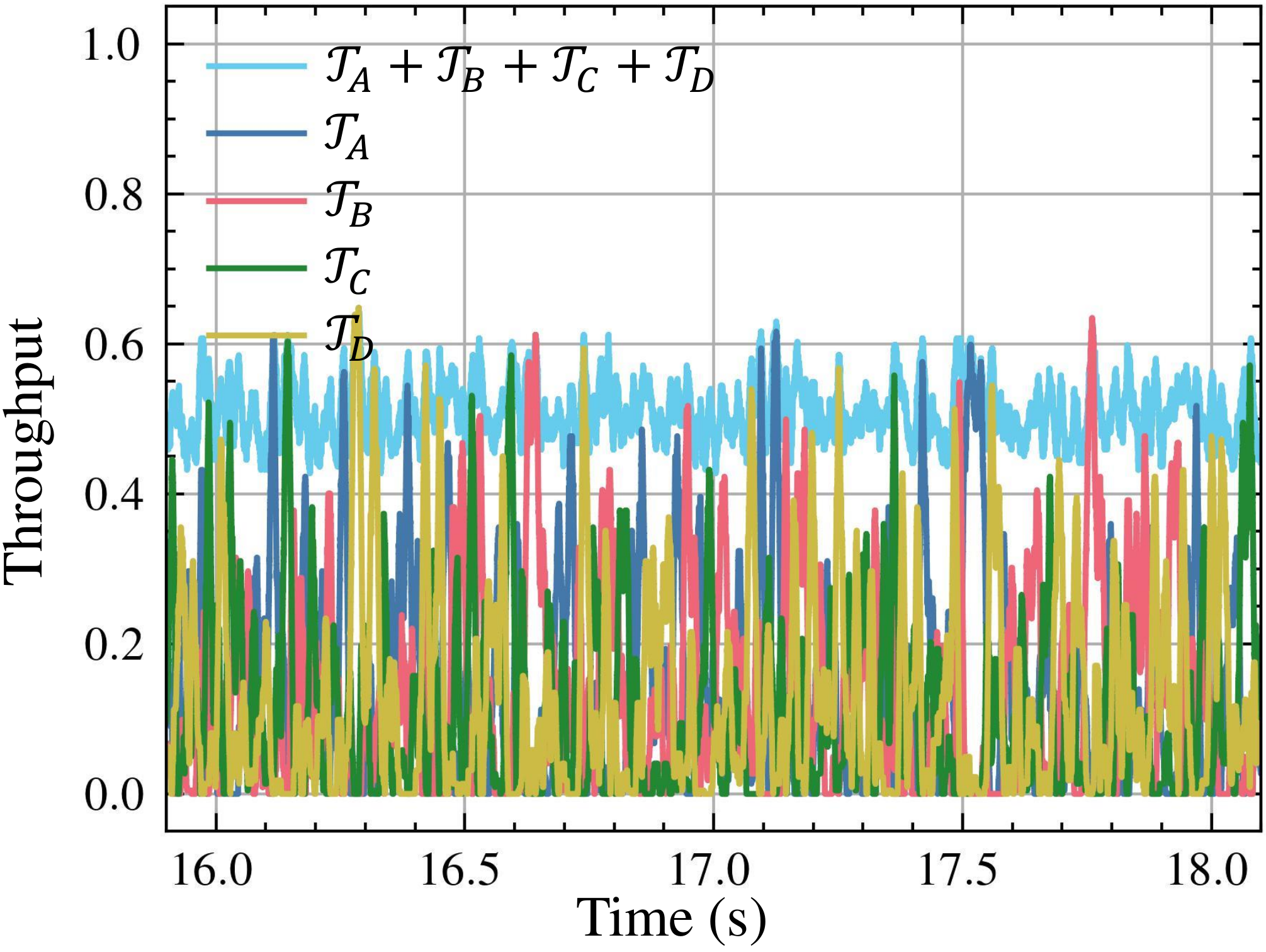}
        \caption{CSMA/CA}
    \end{subfigure}
    \begin{subfigure}[t]{0.245\linewidth}
        \centering
        \includegraphics[width=\linewidth]{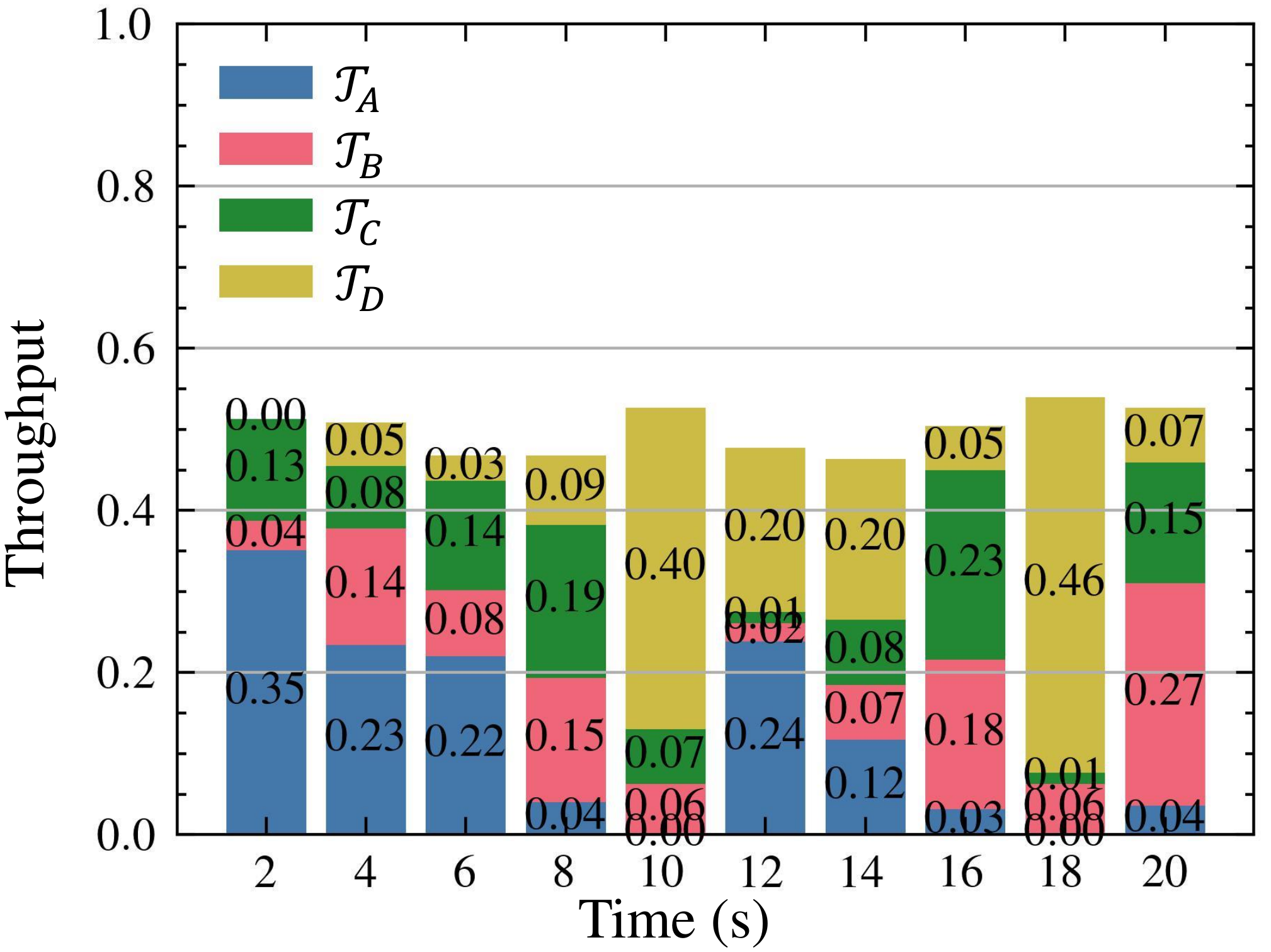}
        \caption{CSMA/CA}
    \end{subfigure}
    \begin{subfigure}[t]{0.245\linewidth}
        \centering
        \includegraphics[width=\linewidth]{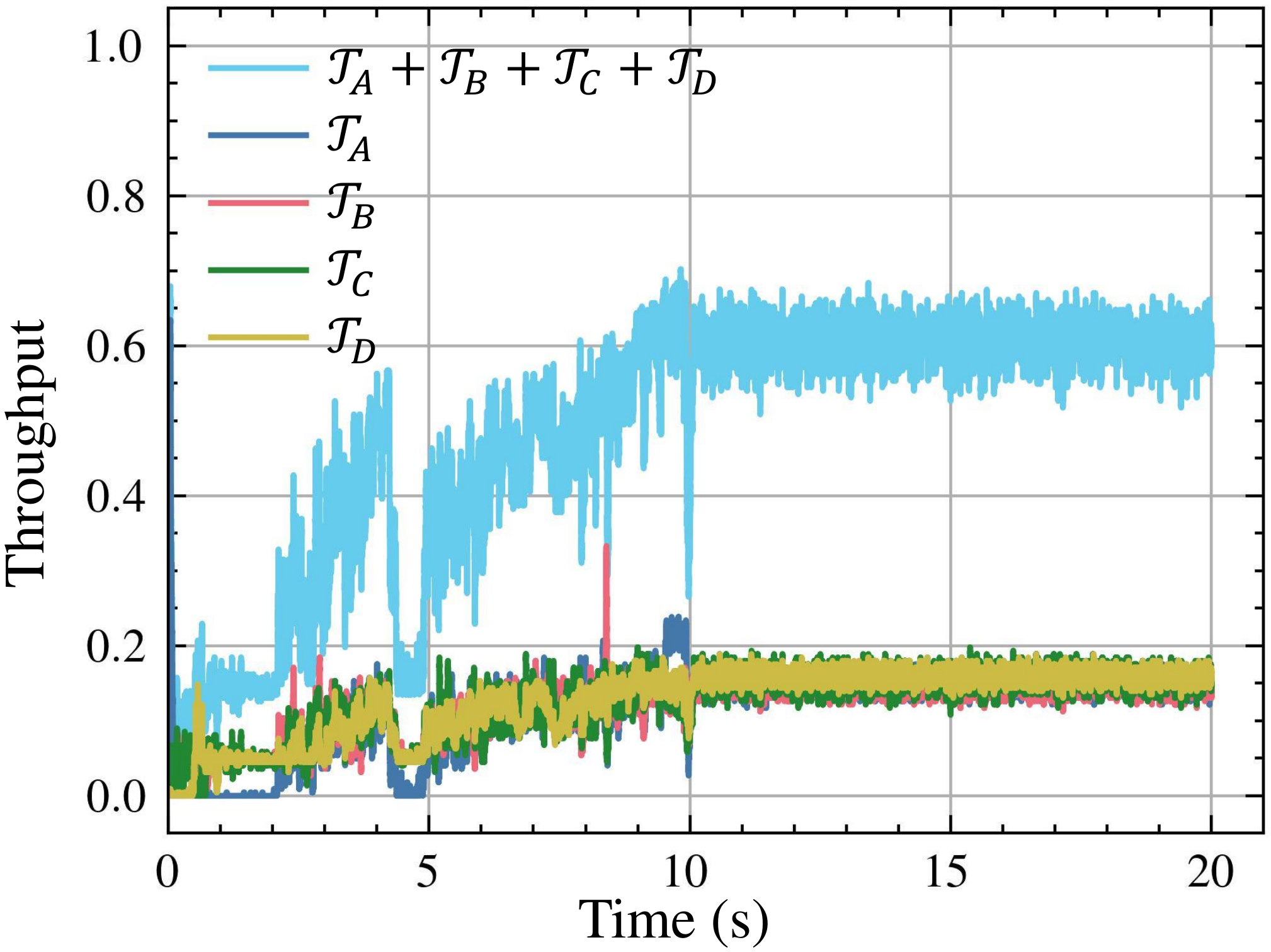}
        \caption{MADRL-HT}
    \end{subfigure}
    \begin{subfigure}[t]{0.245\linewidth}
        \centering
        \includegraphics[width=\linewidth]{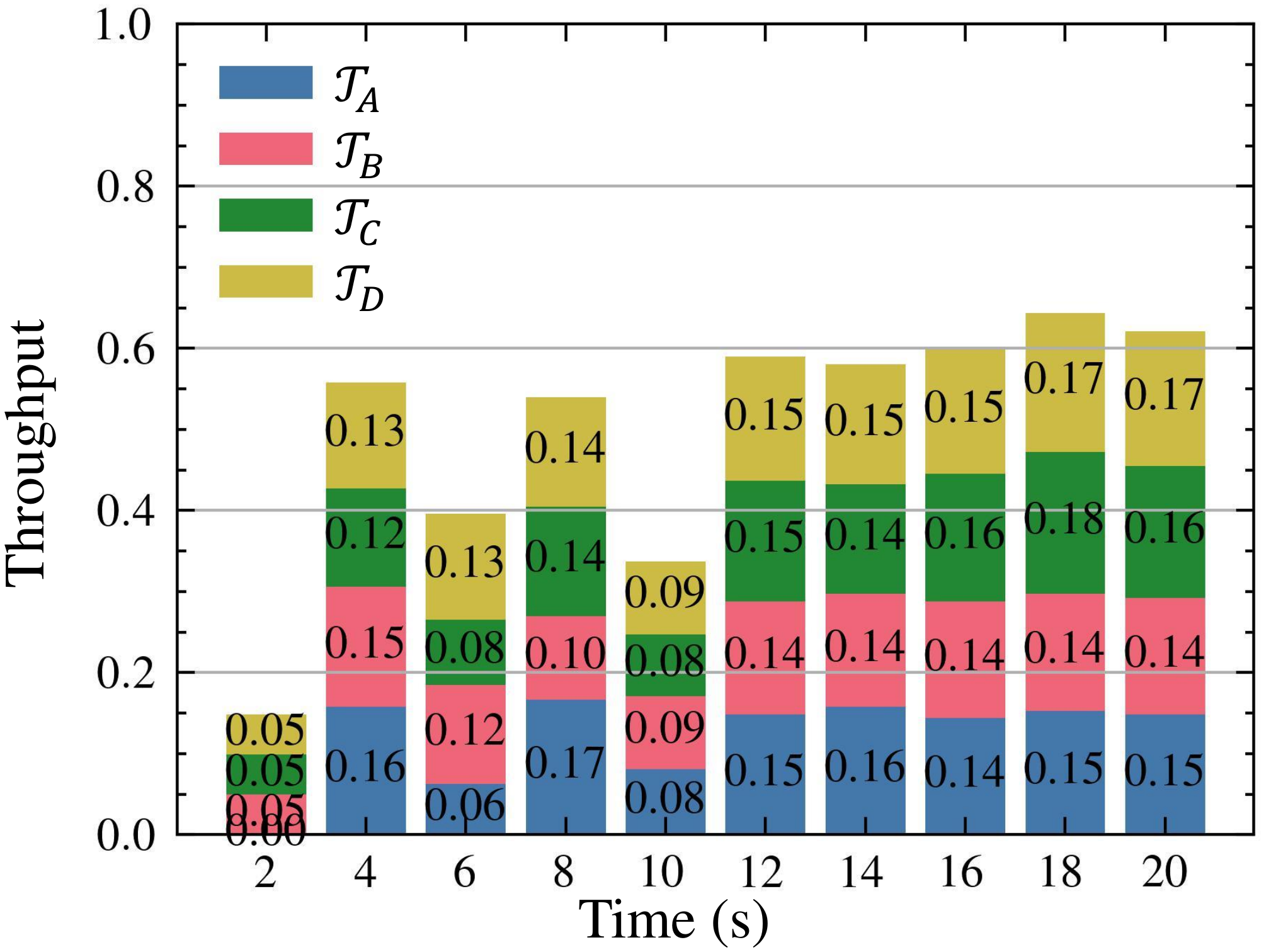}
        \caption{MADRL-HT}
    \end{subfigure}
    \caption{Topo4: the achieved throughput of each terminal in the BSS over the course of training. (a) and (b) are the performance of the CSMA/CA BSS;  (c) and (d) are the performance of the MADRL-HT BSS.}
    \label{fig:S8}
\end{figure*}

\begin{figure*}
    \centering
    \begin{subfigure}[t]{0.245\linewidth}
        \centering
        \includegraphics[width=\linewidth]{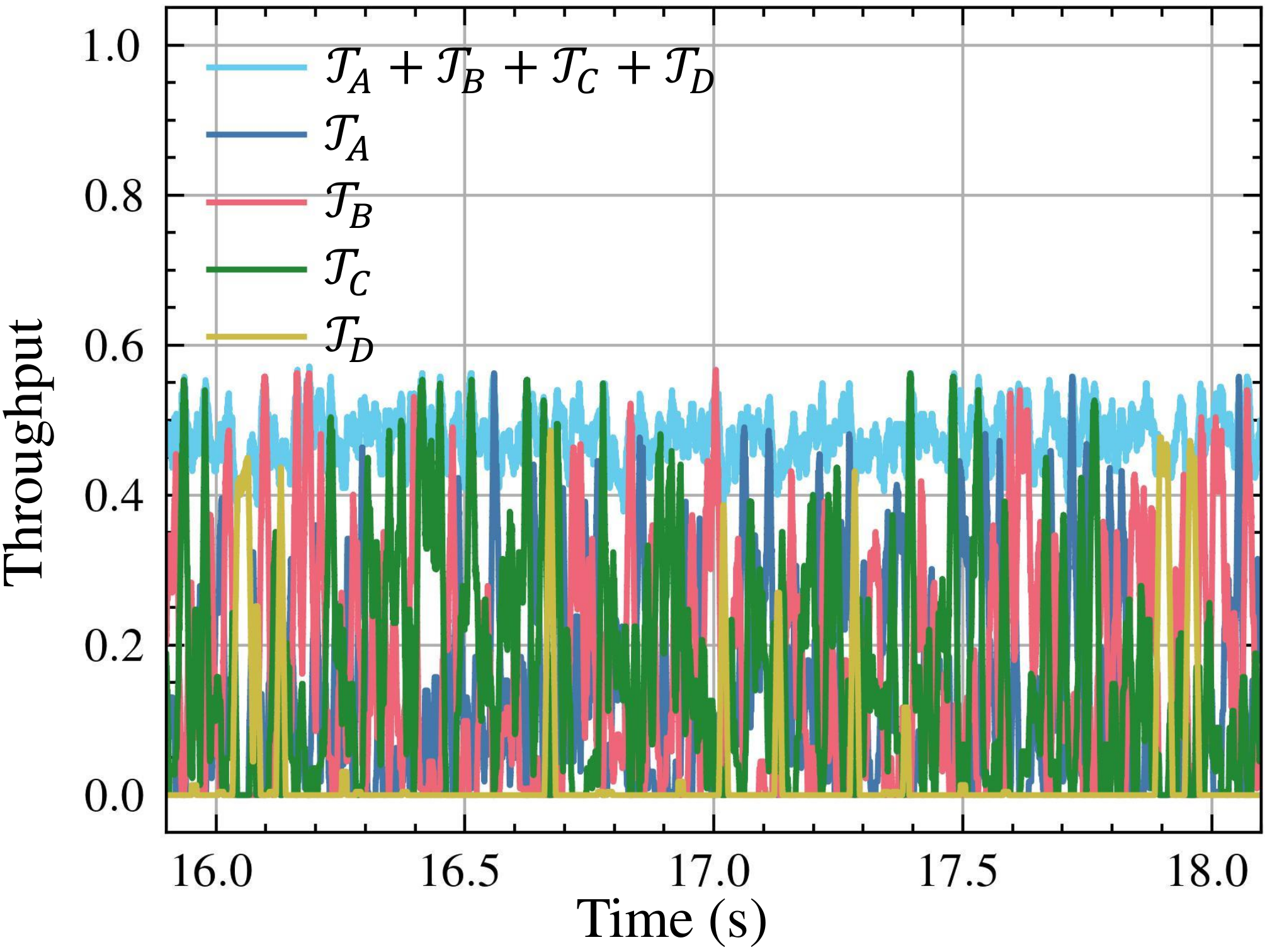}
        \caption{CSMA/CA}
    \end{subfigure}
    \begin{subfigure}[t]{0.245\linewidth}
        \centering
        \includegraphics[width=\linewidth]{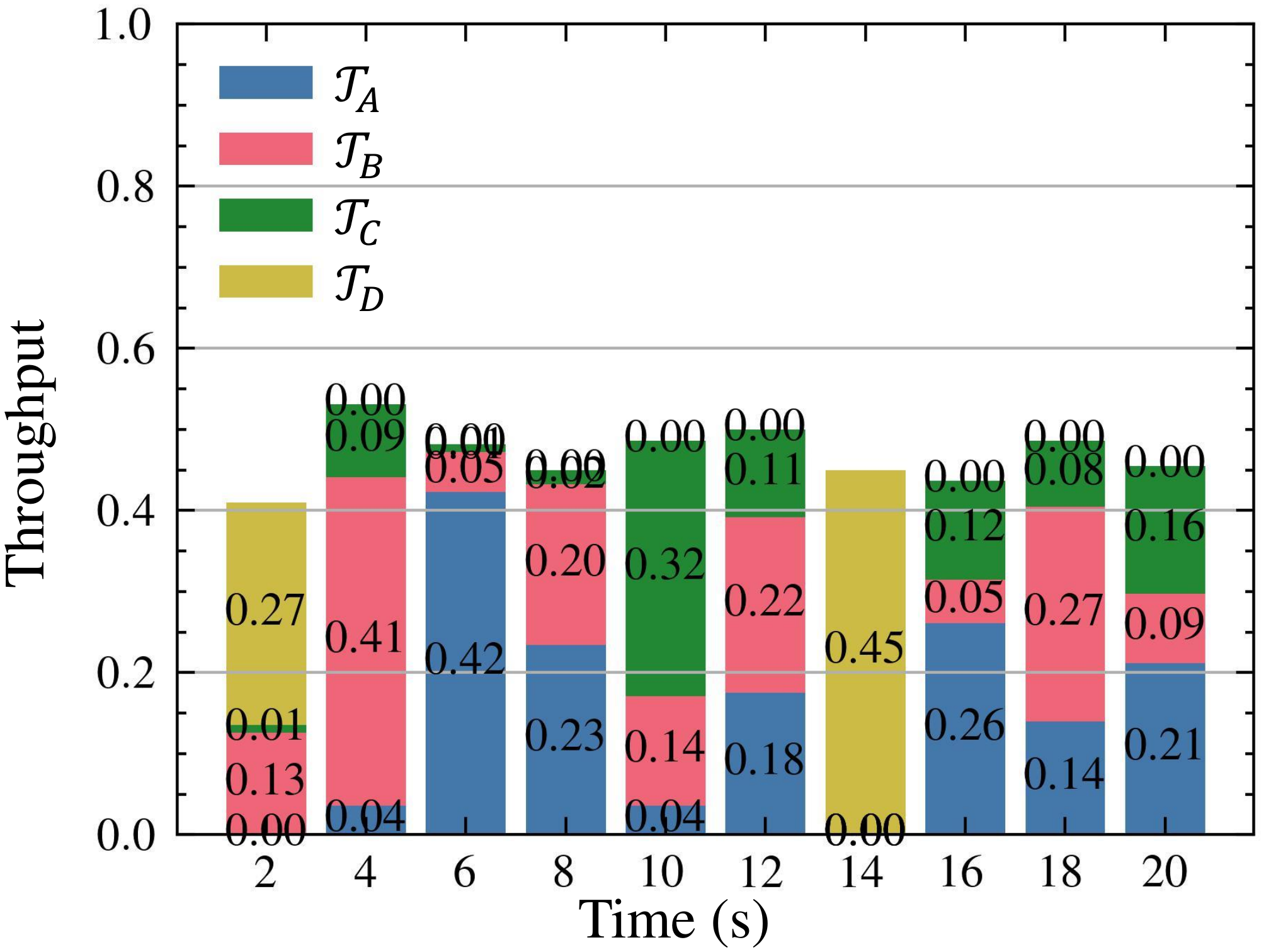}
        \caption{CSMA/CA}
    \end{subfigure}
    \begin{subfigure}[t]{0.245\linewidth}
        \centering
        \includegraphics[width=\linewidth]{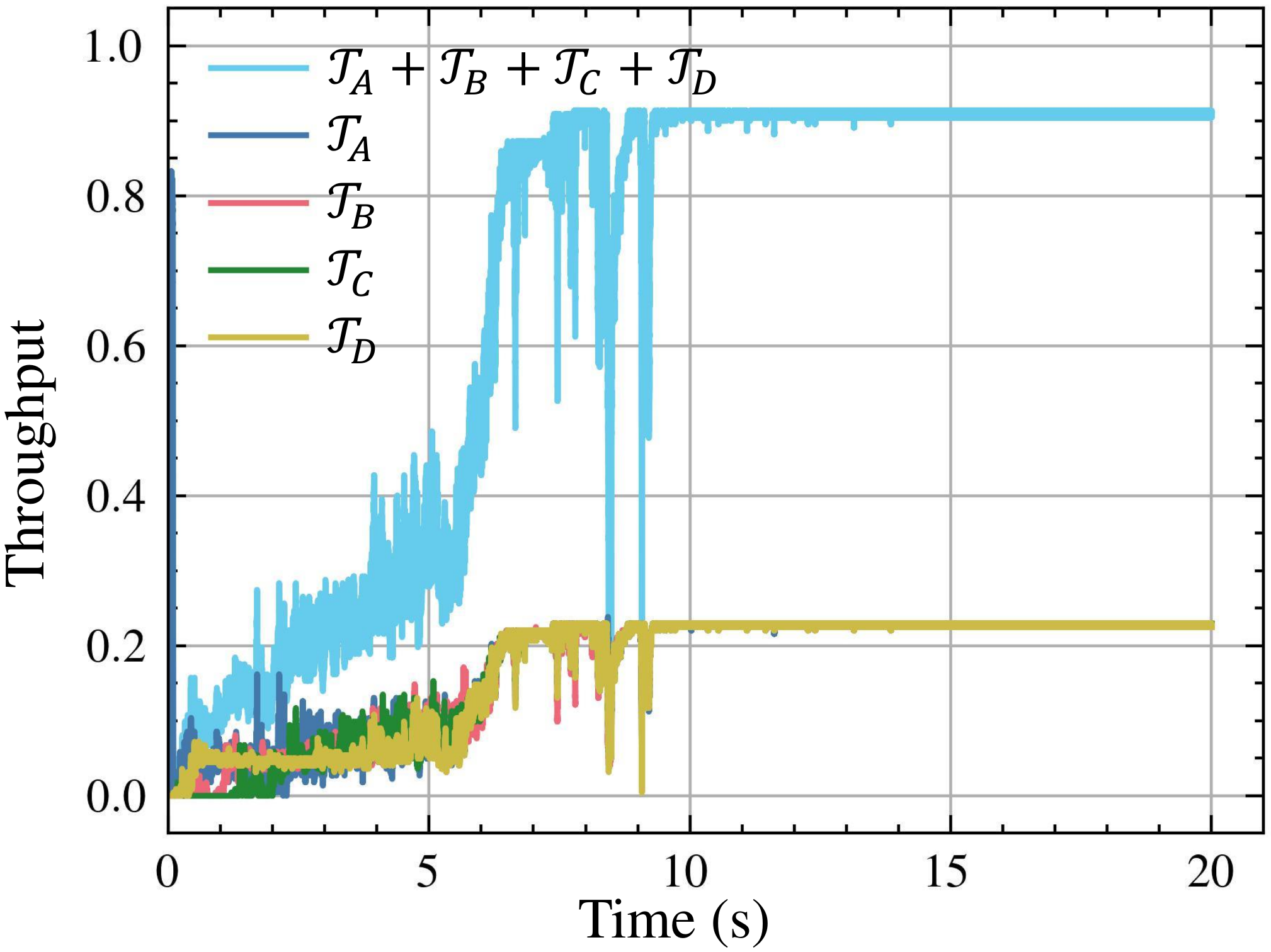}
        \caption{MADRL-HT}
    \end{subfigure}
    \begin{subfigure}[t]{0.245\linewidth}
        \centering
        \includegraphics[width=\linewidth]{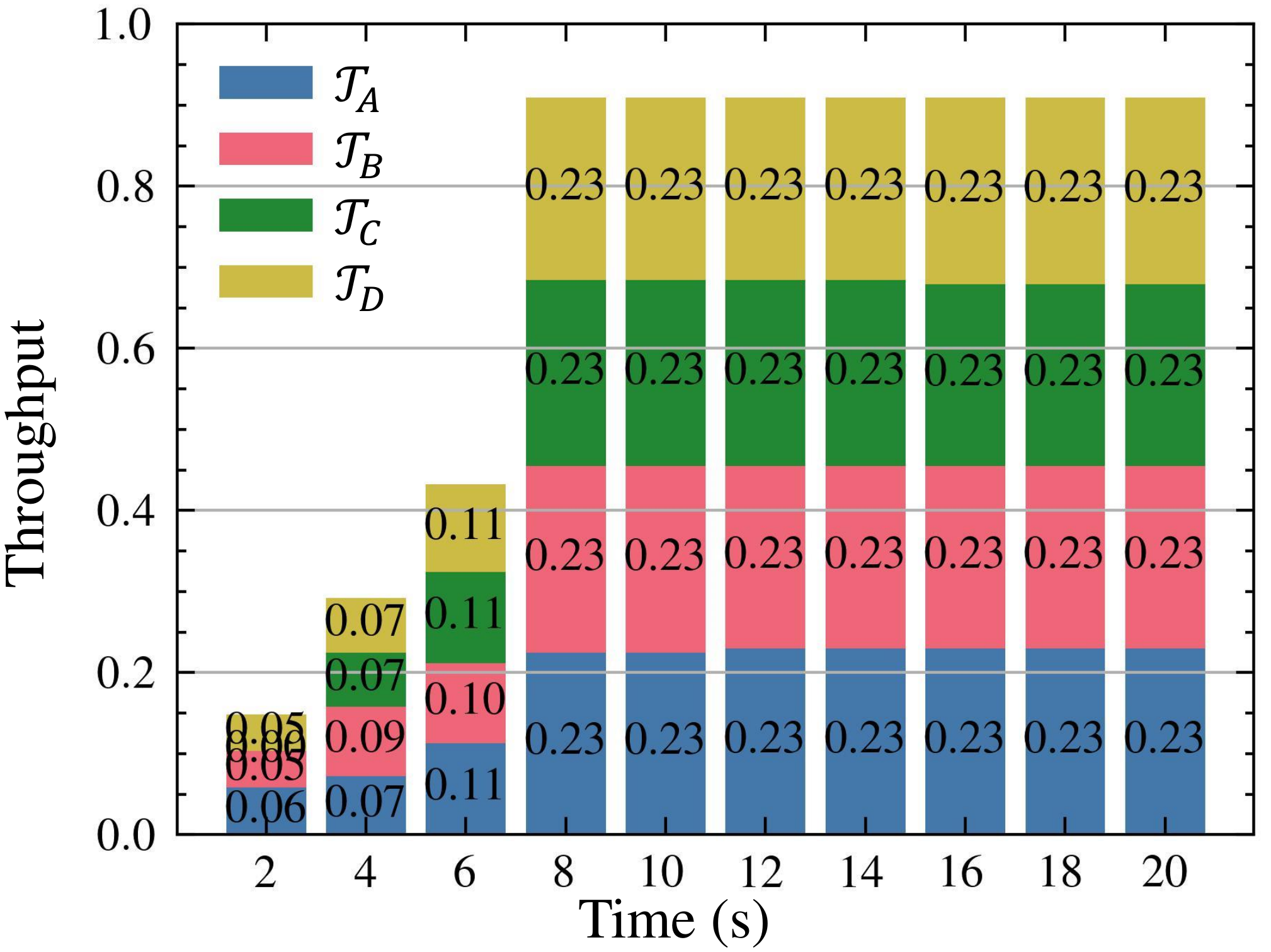}
        \caption{MADRL-HT}
    \end{subfigure}
    \caption{Topo4': the achieved throughput of each terminal in the BSS over the course of training. (a) and (b) are the performance of the CSMA/CA BSS;  (c) and (d) are the performance of the MADRL-HT BSS.}
    \label{fig:S7}
\end{figure*}

{\it 2) Throughput of individual terminals.}
We first study the achieved throughput of individual terminals.
In Topo4, there is no hidden terminal. The achieved throughputs of individual terminals in both BSSs are shown in Fig.~\ref{fig:S8} by curves, i.e., (a) and (c), and histograms i.e., (b) and (d).
Likewise, the throughput of each terminal versus training time in Topo4' (with hidden terminal) is presented in Fig.~\ref{fig:S7}.

We have two main observations from Figures~\ref{fig:S8} and \ref{fig:S7}.
\begin{itemize}
\item For the CSMA/CA BSS, the channel-access policy of each terminal is fixed. The throughput of each terminal is characterized by fluctuations   -- the throughput of individual terminals exhibits an impulse characteristic, and is quite unfair in both Topo4 and Topo4'. 
Comparing Fig~\ref{fig:S8}(b) with Fig~\ref{fig:S7}(b), hidden terminals degrade the BSS throughput and aggravate the unfairness among terminals.
\item For the MADRL-HT BSS, the throughput achieved by individual terminals is almost equal, and the BSS throughput outperforms the CSMA/CA BSS in both Topo4 and Topo4'. An interesting observation is that MADRL-HT learns better in Topo4' than that in Topo4: the convergence is faster and the optimal BSS throughput is achieved.
\end{itemize}

\begin{figure*}
    \centering
    \begin{subfigure}[t]{0.245\linewidth}
        \centering
        \includegraphics[width=\linewidth]{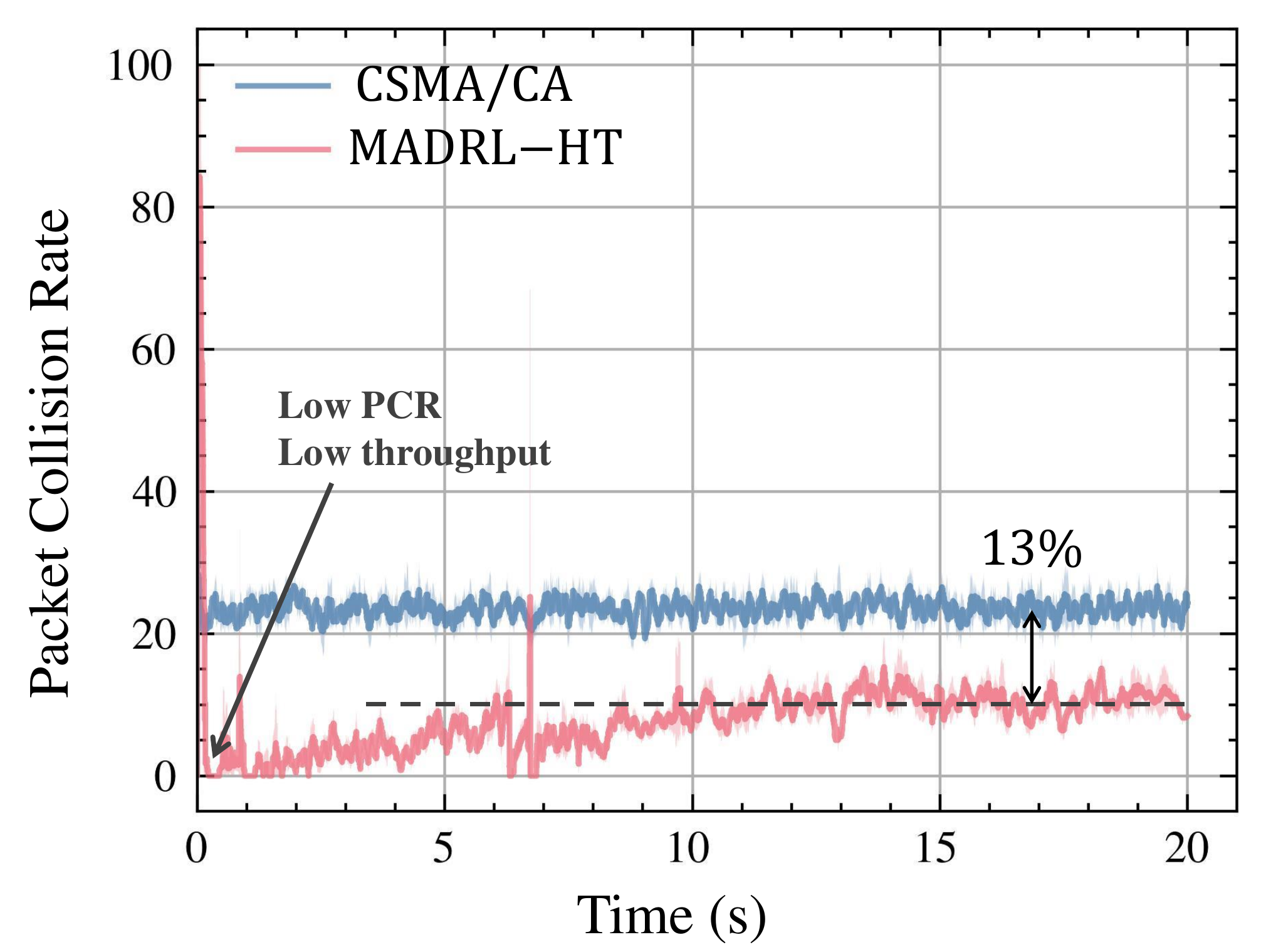}
        \caption{Topo4, PCR}
    \end{subfigure}
    \begin{subfigure}[t]{0.245\linewidth}
        \centering
        \includegraphics[width=\linewidth]{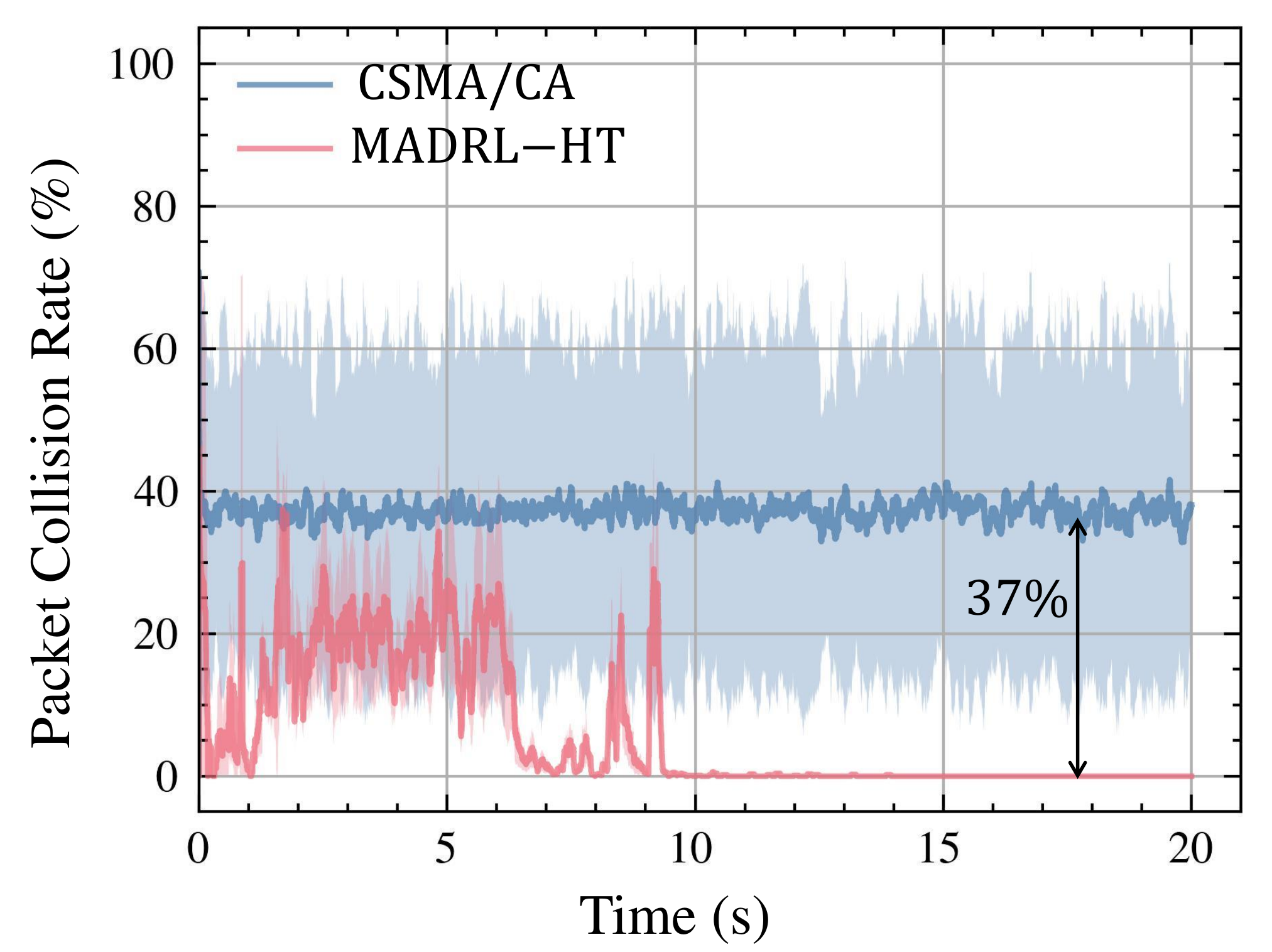}
        \caption{Topo4', PCR}
    \end{subfigure}
    \begin{subfigure}[t]{0.245\linewidth}
        \centering
        \includegraphics[width=\linewidth]{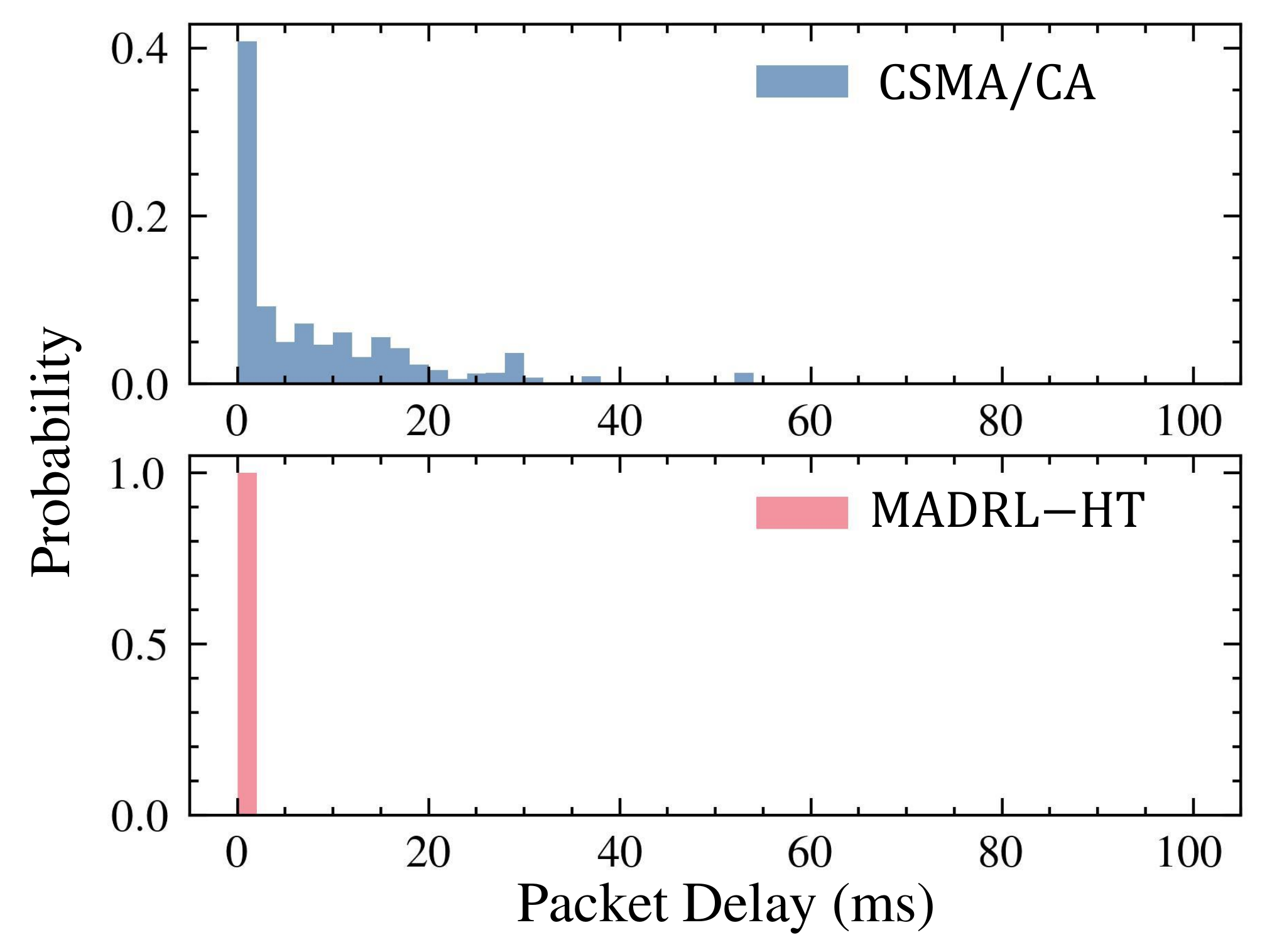}
        \caption{Topo4, packet delay}
    \end{subfigure}
    \begin{subfigure}[t]{0.245\linewidth}
        \centering
        \includegraphics[width=\linewidth]{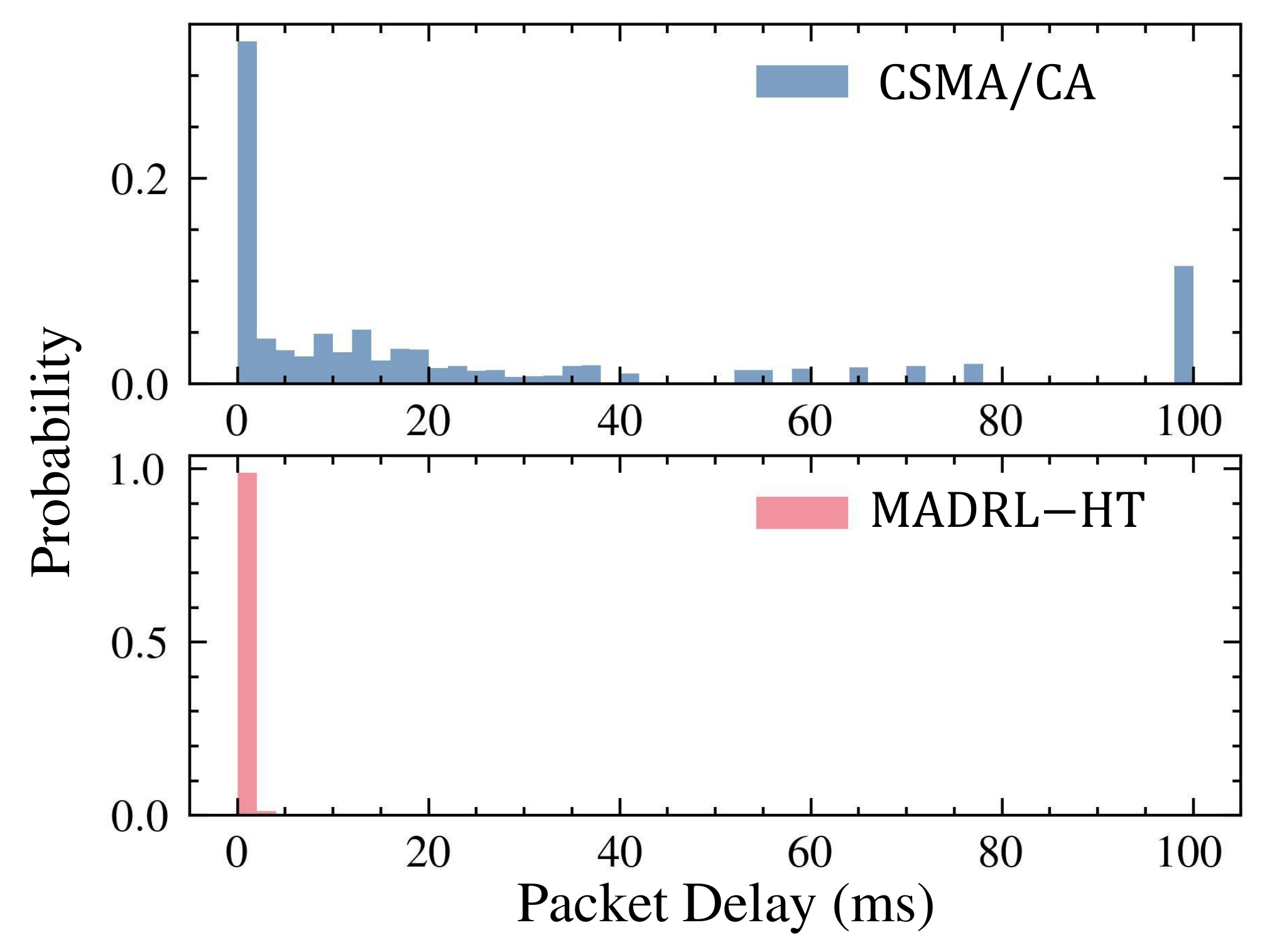}
        \caption{Topo4', packet delay}
    \end{subfigure}
    \caption{The performance of the CSMA/CA BSS and the MADRL-HT BSS: (a-b) the average packet collision rate (PCR), (c-d) the PDF of packet delay, where a packet is dropped if it cannot be transmitted in $100$ms.}
    \label{fig:S10}
\end{figure*}

{\it 3) Packet collision rate (PCR).}
Fig.~\ref{fig:S10}(a-b) presents the PCR for both BSSs, where PCR is defined as the ratio between the number of collisions and the total transmission attempts.
When hidden terminals are presented, the average PCR of the CSMA/CA BSS increases from $23\%$ (in Fig.~\ref{fig:S10}(a)) to $37\%$ (in Fig.~\ref{fig:S10}(b)).
For the MADRL-HT BSS, on the other hand, the PCR reduces very quickly in the initial phase of training (within $0.2$s) in both Topo4 and Topo4'. However, as can be seen from Fig.~\ref{fig:S8}(c) and Fig.~\ref{fig:S7}(c), the system throughput is quite low at that time. This means that the terminals' transmission policies are conservative -- they prefer not to transmit to avoid negative rewards caused by collisions.
As learning proceeds, the terminals start to explore more active policies to reap positive rewards. Although the PCR increases, the system throughput steadily improves. When the learning converges, the average PCR of the MADRL-HT BSS is $13\%$ lower than that of the CSMA/CA BSS in Topo4, and $37\%$ lower in Topo4'.

{\it 4) Delay and jitter.}
The final performance measure we shall evaluate is the packet delay and jitter \cite{jitter} of the two BSSs. For both Topo4 and Topo4', we run the CSMA-CA BSS and the MADRL-HT BSS (using the trained model after $20$s) for $1$s, record the delay of all packets, and plot the probability density function (PDF) of the packet delay in Fig.~\ref{fig:S10}(c-d). 
In particular, a packet will be dropped if it is not transmitted within $100$ms.
In Topo4, the average packet delay of the CSMA/CA BSS is $8.23$ms and the delay jitter  is $10.17$ms.
When it comes to Topo4' with a hidden terminal, the average packet delay increases to $25.06$ms, and the delay jitter increases to $32.93$ms.
In contrast, the average packet delays of the MADRL-HT BSS under Topo4 and Topo4' are only $0.44$ms and $0.21$ms, respectively, while the jitters of the packet delay are $0.22$ms and $0.03$ms, respectively. 
To summarize, with our MADRL-HT solution, the average packet delay is reduced by $94.7\%$ and $99.2\%$, and the delay jitter is reduced by $97.9\%$ and $99.9\%$, in Topo4 and Topo4', respectively.

Overall, our MADRL-HT solution attains excellent performance gains over the legacy CSMA/CA protocol in terms of various QoS metrics.
Nevertheless, it is worth noting that our solution does not always converge to the optimal. Take Topo4' in Fig.~\ref{fig:S6}(e) for example. There is still a $10\%$ gap between the average $\alpha$-fairness achieved by the MADRL-HT BSS and the global optimum. A more general observation is that obtaining a good set of transmission policies is becoming increasingly harder as the number of terminals increases.
In actuality, this phenomenon is not surprising.
More terminals lead to a more intricate transmission history. To exploit useful insights and patterns from history for decision-making, the learning task itself becomes much more complex.
In our MADRL-HT design, the dimension of the input matrix  to an actor is $3\times W$. The number of rows is fixed to $3$ regardless of the number of terminals, thanks to the topological analysis in Section \ref{sec:III}.
The number of columns $W$ (i.e., the look-back window length), however, depends on the number of terminals in the BSS -- a BSS with more terminals requires a larger $W$ to capture enough interactions among terminals to make the right decision.
With the increase in $W$, the actors should be able to extract long-range correlations among the inputs, and this poses greater challenges to state-of-the-art DNNs.
Empirically, with more terminals, the learning is more likely to converge to a local minimum.
The above challenges call for more efficient schemes to compress the columns of the input matrix and extract useful features, which we leave for future work. 

\begin{figure*}
    \centering
    \begin{subfigure}[t]{0.245\linewidth}
        \centering
        \includegraphics[width=\linewidth]{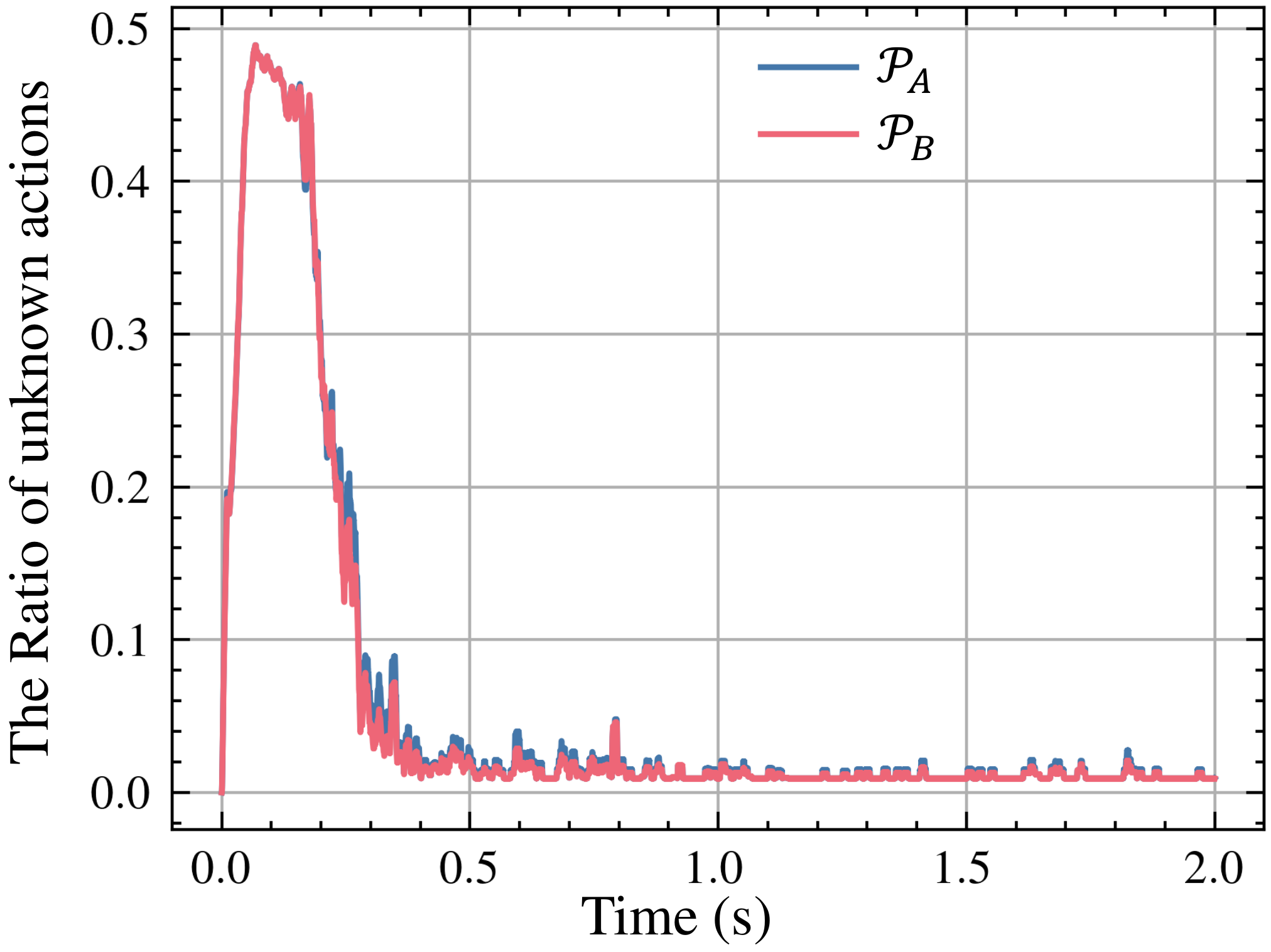}
        \caption{Evaluating unknown actions}
    \end{subfigure}
    \begin{subfigure}[t]{0.245\linewidth}
        \centering
        \includegraphics[width=\linewidth]{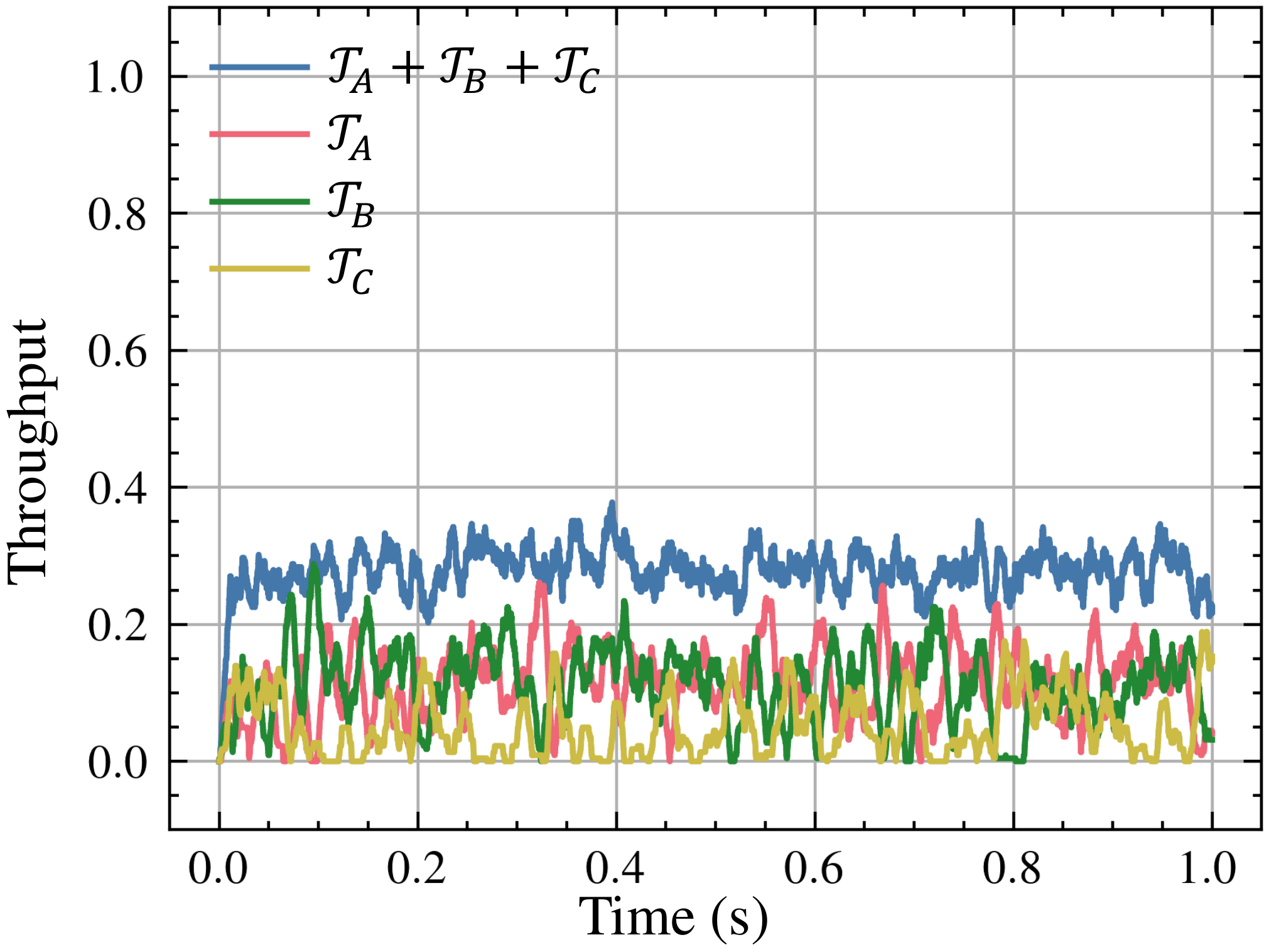}
        \caption{CSMA/CA}
    \end{subfigure}
    \begin{subfigure}[t]{0.245\linewidth}
        \centering
        \includegraphics[width=\linewidth]{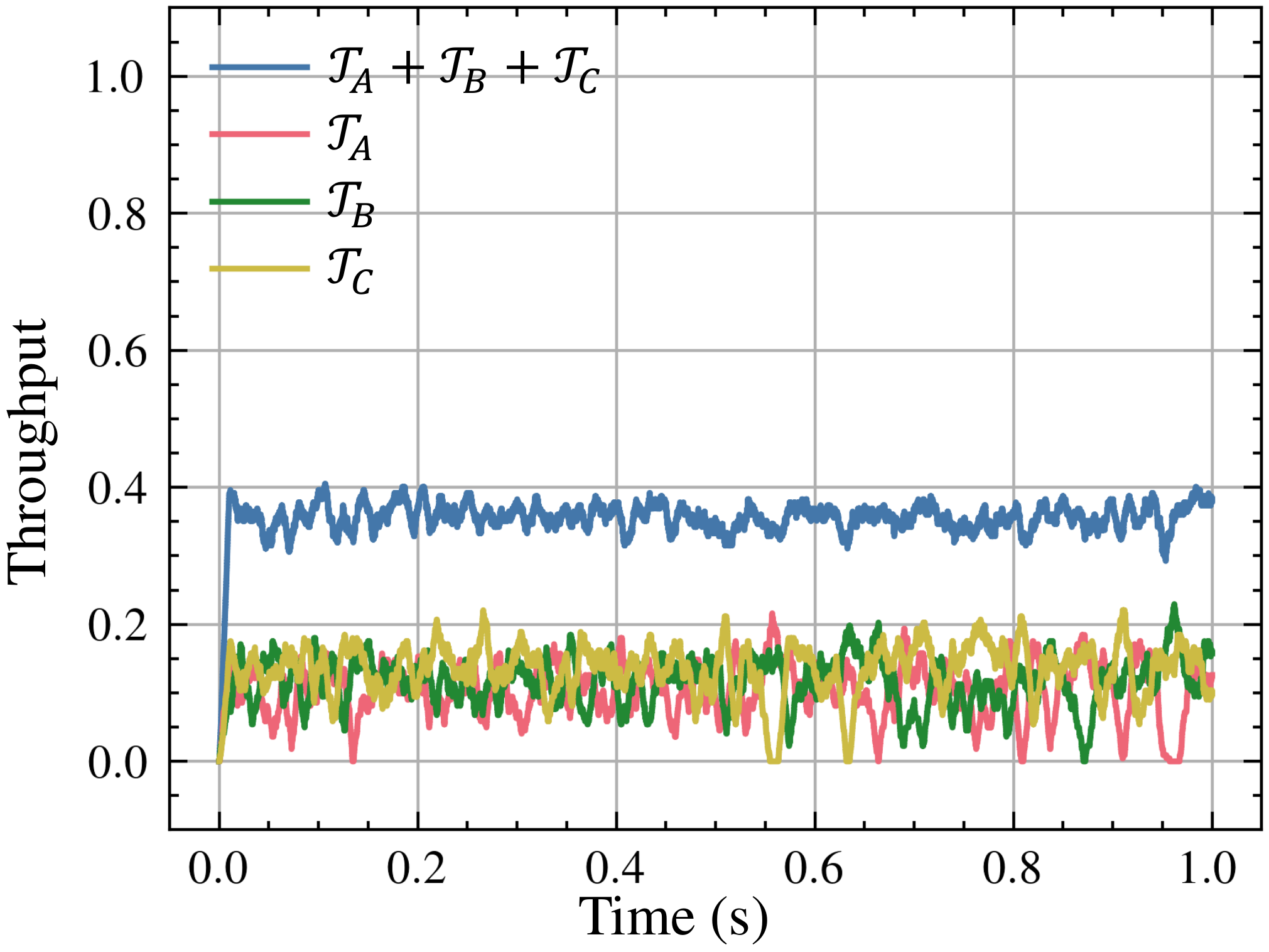}
        \caption{CSMA/CA with RTS/CTS}
    \end{subfigure}
    \begin{subfigure}[t]{0.245\linewidth}
        \centering
        \includegraphics[width=\linewidth]{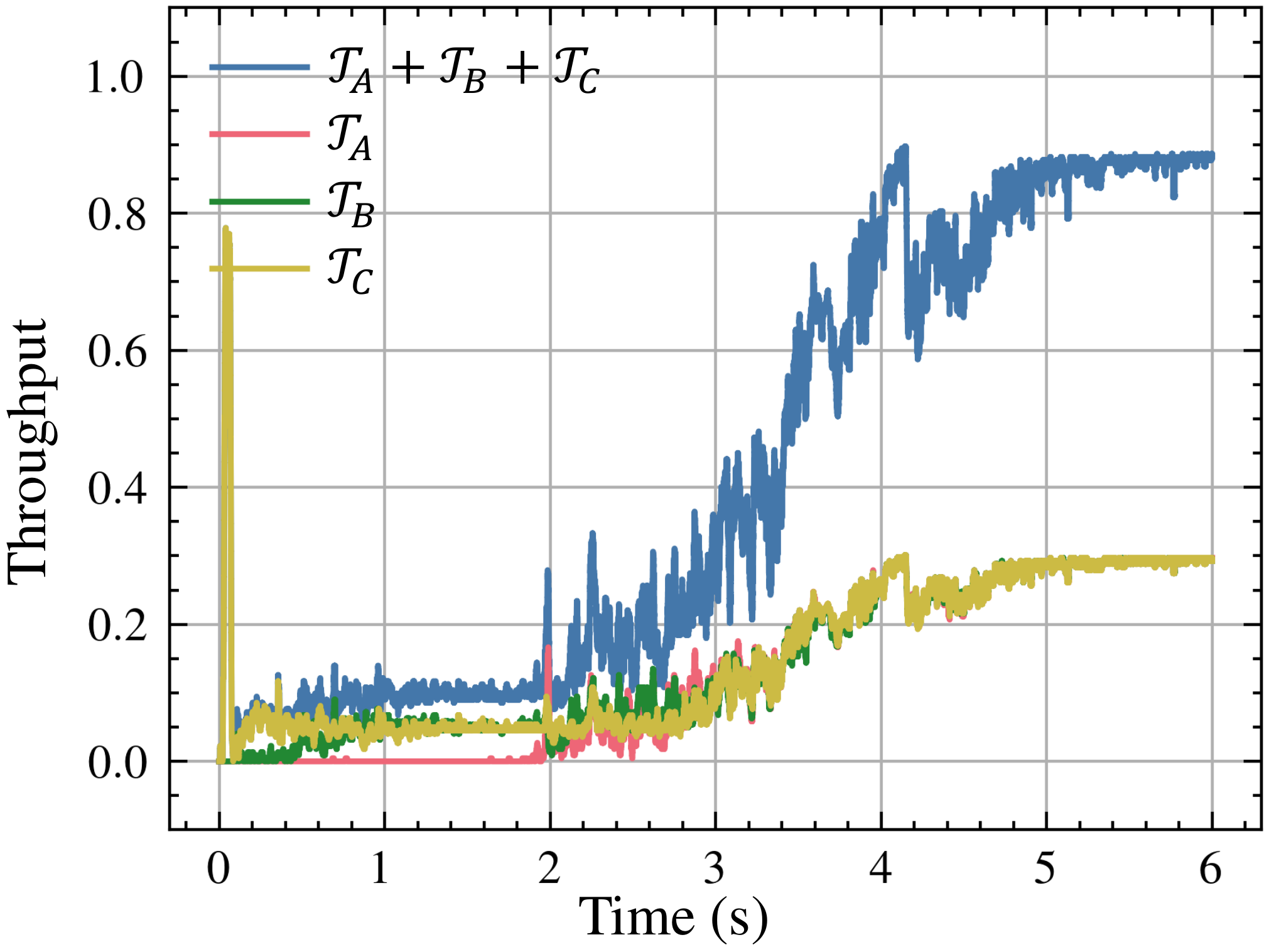}
        \caption{MADRL-HT}
    \end{subfigure}
    \caption{(a) The number of remaining unknown actions in the terminals' observations as a function of the training time in Topo2'. (b-d) A comparison among CSMA/CA, CSMA/CA with RTS/CTS, and MADRL-HT in Topo3’.}
    \label{fig:unk}
\end{figure*}

We specify that the learning performance of MADRL-HT can be reflected by the number of remaining unknown actions in the terminals' observations.
One example is given in Fig.~\ref{fig:unk}(a), where we consider Topo2' and plot the portion of unknown actions in each terminal's observation.
As can be seen, the portion of unknown actions is large at the beginning of training. In this phase, the system throughput is also very poor.
As learning progresses, the number of unknowns in the observations starts to decrease, and finally, vanishes. Correspondingly, the system throughput reaches the global optimum.

Finally, we compare the performance of MADRL-HT with ``CSMA/CA with RTS/CTS''. Under our setup, the packet length is fixed to $5$ slots. This corresponds to a packet of about $640$ bytes, if we consider an OFDM system with $80$ MHz bandwidth, $312.5$ kHz subcarrier spacing, $16$QAM modulation, and $1/2$ channel coding rate.
For RTS/CTS, we assume that both RTS and CTS packets consume only one slot and there is no SIFS between them (i.e., the CTS feedback is instantaneous).
The simulation is performed in Topo3' and the results are shown in Fig.~\ref{fig:unk}(b-d).
As can be seen,
\begin{itemize}
    \item Compared with CSMA/CA, the RTS/CTS mechanism resolves the hidden terminal problem, and hence, improves both the throughput performance and fairness among terminals. Nevertheless, the throughput gain is not significant, because the RTS and CTS packets themselves consume extra channel resources.
    \item Our MADRL-HT scheme achieves the best performance in terms of both throughput and fairness. When the training converges, the throughput and fairness are close to optimal.
\end{itemize}

\section{Conclusion}\label{sec:Conclusion}
This paper developed a new learning paradigm, dubbed MADRL-HT, for AutoCA in the presence of hidden terminals. With MADRL-HT, a group of terminals learns a set of transmission policies autonomously based on their perceptions of the environment and the feedback from the AP, whereby they adapt to each other's transmission behaviors and achieve a fair and high-throughput multiple-access protocol in a distributed fashion.

The challenges of this problem are twofold:
1) The hidden terminal problem: When hidden terminals are present, part of the environment becomes unobservable to the terminals. To achieve a good set of transmission policies, it is essential for the terminals to infer the hidden terminals' behaviors.
2) Non-stationarity: The non-stationarity of the environment is an inherent problem of MARL since agents are environment to each other and their policies evolve continuously over the course of learning. In AutoCA, the environment is the channel state and is governed directly by the transmission behaviors of the terminals. As a result, the environment faced by each terminal is the highly non-stationary behaviors of other terminals, especially in the initial phase of training. Additionally, due to the presence of hidden terminals, terminals also face hidden non-stationarity.

To meet the above challenges, this paper made advances in three main aspects of the learning paradigm for AutoCA. Our main contributions are summarized as follows.
\begin{enumerate}
\item For the first time in the literature, we formulated the AutoCA problem with the practical challenge of hidden terminals and put forth an actor-critic MADRL paradigm to solve it. The superior performance of our MADRL-HT solution was verified by extensive numerical experiments benchmarked against the legacy CSMA/CA protocol in terms of various QoS metrics.
\item To develop a scalable learning paradigm, we drew topological insights, and for each terminal, partitioned the other terminals into two groups according to the number of communication hops. In doing so, from a terminal's point of view, all other terminals can be viewed as two consolidated terminals regardless of their numbers. More broadly, the idea of terminal clustering based on the number of communication hops is essential to devise a scalable and compact learning paradigm, and potentially shed new light on efficient learning algorithm designs in large-scale networks.
\item To compensate for the partial observability and alleviate the hidden non-stationarity, we proposed a look-back mechanism to revise the carrier-sensed channel state and infer the hidden terminals' behaviors from the ACK/NACK feedback of the AP. The look-back mechanism can be easily extended to other wireless networking problems with hidden terminals.
\item To maximize the network throughput while guaranteeing fairness, we devised a new reward function that instructs the terminals based on their relative number of successful transmissions over a look-back window of time. Our reward focuses more on the relative performances of the terminals; a positive global reward can be gained as long as their successful transmissions do not differ too much. As a result, the learning of terminals proceeds in a more progressive fashion. We empirically found that the window-based reward is particularly useful at the early stages of training when the transmission policies vary drastically.
\end{enumerate}
 
Addressing the hidden terminal problem is an important step in developing practical learning paradigms for AutoCA, and has a significant impact on general random access protocols for wireless ad-hoc networks.
Along this direction, the scope of our future work concerns larger-scale networks, including the single BSS setup with tens of terminals and the multi-BSS setup with multiple channels. The challenge in the former scenario lies in designing more efficient schemes to compress the transmission history during the look-back window, as discussed in Section~\ref{sec:V}; the challenge in the latter scenario, on the other hand, lies in channel management across BSSs.

In the big picture, our vision is to achieve a fully intelligent wireless network, wherein the terminals are able to make independent decisions and network with each other, with and without a central controller.
As a first step, this paper considers the Wi-Fi scenario with a central controller, i.e., the AP.
This setup simplifies the general problem as the AP can provide global information (in our case, the advantage function given by the critic network) to assist the learning of individual terminals.
Hopefully, our study in this paper can provide general insights for achieving the big picture of wireless intelligent networks, wherein autonomous MAC algorithms are indispensable parts.

\bibliographystyle{IEEEtran}
\bibliography{References}

\end{document}